\documentclass{article}
\usepackage{PRIMEarxiv}
\usepackage[utf8]{inputenc} 
\usepackage[T1]{fontenc}    
\usepackage{url}            
\usepackage{booktabs}       
\usepackage{amsfonts}       
\usepackage{nicefrac}       
\usepackage{microtype}      
\usepackage{lipsum}
\usepackage{fancyhdr}       
\usepackage{graphicx}       
\usepackage{amssymb}
\usepackage{subcaption}
\usepackage[dvipsnames]{xcolor}
\usepackage{array}
\usepackage{chemformula}
\usepackage[backend=biber, style=ieee]{biblatex}
\graphicspath{{media/}}     

\title{Interpretable Multi-Source Data Fusion Through Latent Variable Gaussian Process}

\author{
  Sandipp Krishnan Ravi$^{1,*}$ \\
  Probabilistic Design and Material Informatics Group \\
  GE Aerospace Research \\
  Niskayuna, NY \\
  \texttt{sandippkrishnan.ravi@ge.com} \\
   \And
  Yigitcan Comlek$^1$ \\
  Integrated Design Automation Laboratory \\
  Northwestern University \\
  Evanston, IL \\
 \texttt{yigitcancomlek2024@u.northwestern.edu} \\
    \And
  Arjun Pathak \\
  Metallurgy Group \\
  GE Aerospace Research \\
  Niskayuna, NY \\
   \And
  Vipul Gupta \\
  Metallurgy Group \\
  GE Aerospace Research \\
  Niskayuna, NY \\
   \And
  Rajnikant Umretiya \\
  Metallurgy Group \\
  GE Vernova Research \\
  Niskayuna, NY \\
   \And
  Andrew Hoffman \\
  Metallurgy Group \\
  GE Vernova Research \\
  Niskayuna, NY \\
   \And
  Ghanshyam Pilania \\
  Probabilistic Design and Material Informatics Group \\
  GE Aeropace Research \\
  Niskayuna, NY \\
   \And
  Piyush Pandita \\
  Probabilistic Design and Material Informatics Group \\
  GE Aerospace Research \\
  Niskayuna, NY \\
   \And
  Sayan Ghosh \\
  Probabilistic Design and Material Informatics Group \\
  GE Aerospace Research \\
  Niskayuna, NY \\
   \And
  Nathaniel Mckeever \\
  Materials Characterization \\
  GE Aerospace Research \\
  Niskayuna, NY \\
   \And
  Wei Chen$^*$ \\
  Integrated Design Automation Laboratory \\
  Northwestern University \\
  Evanston, IL \\
 \texttt{weichen@northwestern.edu} \\
    \And
  Liping Wang\\
  Probabilistic Design and Material Informatics Group \\
  GE Aerospace  Research \\
  Niskayuna, NY \\
   \And
}

\begin{document}
\maketitle

\def\thefootnote{1}\footnotetext{Equal Contribution}\def\thefootnote{\arabic{footnote}}
\def\thefootnote{*}\footnotetext{Corresponding Author}\def\thefootnote{\arabic{footnote}}

\begin{abstract}
With the advent of artificial intelligence and machine learning, various domains of science and engineering communities have leveraged data-driven surrogates to model complex systems through fusing numerous sources of information (data) from published papers, patents, open repositories, or other resources. However, not much attention has been paid to the differences in quality and comprehensiveness of the known and unknown underlying physical parameters of the information sources, which could have downstream implications during system optimization. Additionally, existing methods cannot fuse multi-source data into a single predictive model. Towards resolving this issue, a multi-source data fusion framework based on Latent Variable Gaussian Process (LVGP) is proposed. The individual data sources are tagged as a characteristic categorical variable that are mapped into a physically interpretable latent space, allowing the development of source-aware data fusion modeling. Additionally, a dissimilarity metric based on the latent variables of LVGP is introduced to study and understand the differences in the sources of data. The proposed approach is demonstrated on and analyzed through two mathematical and two materials science case studies. From the case studies, it is observed that compared to using single-source and source unaware machine learning models, the proposed multi-source data fusion framework can provide better predictions for sparse-data problems.

\end{abstract}


\section{Introduction}

Numerous domains of engineering have benefited from the advent of artificial intelligence (AI) and machine learning (ML). AI$\backslash$ML frameworks, mostly in the form of predictive models, play a pivotal role in representing complex engineering systems and analyzing intricate relationships within data \cite{nti2022applications}. Their utilization can be further extended to optimizing designs, predicting system behavior, and informing downstream decision-making tasks. Essentially, AI$\backslash$ML frameworks drive innovation across diverse engineering disciplines. With the increase demand and availability of information (data), an emerging area of research for the AI$\backslash$ML community is the fusion of data collected from numerous sources into modeling \cite{meng2020survey}. This phenomenon of forming a collective intelligence through numerous sources of information is also known as data fusion. The concept of data fusion aims to provide a unified point of view regarding a system of interest through integration of information available from different sources \cite{bleiholder2009data}. In the context of data fusion, source is defined as a set of data points or a data generating process with different mapping behavior between the fixed set of inputs and outputs. In most cases, data from different sources are collected and a single ML model that aims to capture the relationship between the design variables and the output of interest is built without considering the possible underlying influence of the information sources. In other words, due to the diverse nature of data collection, each information source could contain different underlying procedures and processes that are not reported and incorporated into the ML modeling. Noteworthy and crucial applications of data fusion, along with the challenges with  the incorporation of information sources are provided in previous works \cite{zhou2019information, yousefpour2024gp+, eweis2022data, foumani2023multi, zanjani2024safeguarding,wang2024deep,yao2023ensemble,zhang2021multi}. The sources of variability between the information sources encompass a wide range of factors, including but not limited to measurement errors, confidential information, and various other contributors. Therefore, a major challenge in data fusion exists for the incorporation of the details and unknown parameters associated with each information source.

In the context of materials engineering, the impact of AI$\backslash$ML and data fusion frameworks has been remarkable \cite{ravi2023uncertainty, ravi2022data, ravi2023elucidating, batra2019multifidelity, pilania2017multi}. These data-driven frameworks enabled rapid screening of vast materials and identification of novel combinations and formulations for the design and discovery of superior materials with desired properties. However, even with such proliferated data-driven material discovery efforts, the availability and quality of data remains an important challenge, especially for system level experimental materials design and discovery applications \cite{himanen2019data}. The challenge of experimental data availability and quality is primarily rooted in the diverse nature of experimental materials data. Typically, the first step towards optimizing a material system for a given functionality starts by accruing data from numerous resources including research literature (papers, patents) and open-source data repositories \cite{blaiszik2016materials,puchala2016materials,gong2022repository,kirklin2015open,jain2013commentary} to form a unified view and understanding for the material through data fusion. On the other hand, unlike computational modeling of material behavior, there exists no particular standardization for manufacturing and experimentally synthesizing material systems. Furthermore, issues related to data quality persist due to inconsistencies in experimental methodologies, measurement errors, lack of standardized reporting practices, and confidentiality \cite{sambasivan2021everyone, liang2022advances}. Therefore, this situation leads to an unavoidable data quality issue that has a severe impact on subsequent data-driven modeling and optimization of these material systems.

To tackle the aforementioned challenges, the need for a multi-source data fusion that considers source variabilities becomes evident. In this work, a Gaussian Process (GP) based multi-source data fusion modeling through Latent Variable Gaussian Process (LVGP) \cite{zhang2020latent} is proposed to fuse numerous information (data) sources together to obtain a single representative interpretable prediction model. LVGP is a novel ML model that allows incorporation of qualitative (categorical) variables in addition to quantitative (continuous) variables into surrogate modeling. Throughout this research, in addition to incorporating engineering system parameters as numerical (and categorical if there are any), LVGP is used to incorporate each data source as a categorical variable into the data fusion modeling. The categorical source variables are incorporated into the data fusion model by implicitly mapping each variable onto a physically-meaningful low-dimensional numerical latent space through statistical inference techniques. Here, the term “physically-meaningful” can be associated with explaining the cause-effect relationships between the information sources and the system response. Therefore, the categorical representation of the information sources can indirectly account for the variability that the sources bring through known and unknown underlying processing variables, which in return provides interpretability regarding the information sources and their relationships on the system of interest. Furthermore, through extraction of meaningful spatial patterns from the latent space, data availability and quality challenges can be mitigating by carefully selecting which sources to be used for modeling. As a result, the innovation contributions of this work are as follows: (1) a single source-aware multi-source data fusion model, (2) interpretable latent space that explains the relationships between sources, (3) a dissimilarity metric on the latent space that can be used for targeted source selection for improved data fusion modeling, and (4) application of the proposed method on real engineering problems.


The outline of the paper is as follows: In Section 2, descriptions of the ML models that are used for data fusion, GP and LVGP, are given and multi-source data fusion through LVGP is introduced. Section 3 demonstrates two exemplary mathematical representative problems to highlight the advantages of using LVGP for multi-source information modeling. In Sections 4 and 5, the proposed multi-source model is applied on two alloy systems (FeCrAl and SmCoFe) to illustrate its benefits for material science research. Finally, the work is concluded through Section 6.

\section{Methodology}
\subsection{Review of Gaussian Process Regression}
With its powerful and versatile capabilities of capturing nonlinear responses and providing fast and accurate predictive modeling with uncertainty quantification, Gaussian processes (GPs) are well-known metamodels that have been implemented in many engineering applications \cite{williams2006gaussian}. GPs are nonparametric regression models and are based upon Bayesian statistics in combination with numerous kernel methods to provide data-driven probabilistic predictions. Overall, GPs provide an immensely powerful tool for engineers seeking to build fast and accurate predictive models while quantifying and managing uncertainty. 

Considering a typical quantitative (numerical or continuous) m-dimensional input space with $x=[x_1,x_2,...,x_m]$ with a response of interest, $y(x)$, a typical single response Gaussian Process model can be described using Equation (1),

\begin{equation}\label{eq:gp}
        y(x)=\mu +K_x
\end{equation}

which consists of prior constant mean, $\mu$, that describes the mean response at any given point ($x$), and a zero-mean Gaussian Process with a covariance function, $K_x$, which determines the relationship or the correlation between variables in the model. The covariance function $K_x$ can be further extended to $K(x,x^{'})=\sigma^2 \cdot c(x,x^{'})$ where the $\sigma^2$ represents the prior variance of the GP model and $c(x,x^{'})$ describes the correlation function between each point in the model through the specified correlation function. For the modeling of the GP model,  the implemented Gaussian kernel is defined as

\begin{equation}\label{eq:corr1}
        c(x,x^{'}) = exp(-\sum_{i =1}^{m} (\phi_{i}(x_{i}-x_{i}^{'})^2) )
\end{equation}

where $\phi$ describes the scale between input variables. The selected kernel specifically describes how similar or correlated the function values are at different input points. For instance, when input variables are spatially closer to each other, their covariance results in high values, indicating a strong correlation between the variables. Conversely, when they are far apart, the covariance decreases, signifying a weaker correlation. The GP model parameters, $\mu$, $\sigma^2$, and $\phi$ are estimated through the maximum likelihood estimation (MLE).

Although GPs are powerful ML models for numerous engineering applications, they tend to perform poorly when qualitative (categorical) variables are present in the input design space. The main challenge arises from the absence of a well-defined distance metric that differentiates qualitative variables. To overcome this vital challenge, Latent Variable Gaussian Process (LVGP) has been developed \cite{zhang2020latent}. The details of the LVGP model are given in the next section.

\subsection{Review of Latent Variable Gaussian Process}

Due to the nature of the GP correlation functions, it is not plausible to directly include qualitative variables into the modelling since the differences between qualitative variables become unclear without a well-defined distance metric. Recently, Latent Variable Gaussian Process (LVGP) has been developed to integrate qualitative variables into GP modeling through quantitative latent variable representation of these variables to address this difficulty \cite{zhang2020latent}. Many prior works have demonstrated the remarkable modeling capabilities of LVGP for different engineering applications and techniques with mixed-variable (continuous and categorical) domains including but not limited to adaptive materials design \cite{iyer2020data,wang2020featureless,zhang2020bayesian,comlek2023rapid,wang2021data}, non-hierarchical interpretable multi-fidelity modeling \cite{CHEN2024116773}, and mixed-variable (continuous and categorical) global sensitivity analysis \cite{comlek2023mixed}.

To briefly review the method, consider a new input space, $\mathbf{b} = [\mathbf{x^T},\mathbf{t^T}]$ where both quantitative, $\mathbf{x}=[x_1,x_2,..x_m] \epsilon R^m$, and qualitative $\mathbf{t}=[t_1,t_2,...,t_q] \epsilon R^q$ input variables are included in this new domain. Each qualitative variable can have $j$ number of options (levels), i.e $t_i = [l_1(t_i),l_2(t_i),...,l_j(t_i)]$ where $i = 1,2,...,q$. For a given response of interest, $y(b)$, the influence of each qualitative variable can be represented by an underlying quantitative (numerical) space, $v_{t_i}(l_j) = [v_1(l_1),v_2(l_1),...,v_k(l_j)] \epsilon R^k$. This underlying quantitative space can possibly be undiscovered, unknown, complex, or high-dimensional. As a result, the main idea of LVGP lies in learning a low-dimensional  quantitative latent space to approximate the original underlying space through statistical inference using the available input-output data. Specifically, it is assumed that this low dimensional space can capture the joint effects of these underlying qualitative variables on the response of interest according to sufficient dimension reduction arguments \cite{li1991sliced,cook2005sufficient}. As such, according to the authors of the original LVGP paper \cite{zhang2020latent}, a two-dimensional latent variable vector is sufficient to express the influence of the qualitative variables on the response. Therefore, qualitative variable, $t_i$ can be represented as $\mathbf{z}(t_i) = [z_1(l_1),z_2(l_1),z_1(l_2),z_2(l_2),...,z_1(l_j),z_2(l_j)]$, where each level $l_j(t_i)$ is represented with a 2D latent vector $[z_1(l_j),z_2(l_j)]$. Thus, the previously defined mixed-variable input space, $\mathbf{b} = [\mathbf{x}^{T},\mathbf{t}^{T}]$, then becomes $\mathbf{w} = [\mathbf{x}^{T},\mathbf{z(t)}{^T}] \epsilon R^{m + 2 \times q}$. With the newly defined input space, the GP correlation function (Equation \ref{eq:corr1}) is updated as,

\begin{equation}\label{eq:corr2}
        c(w,w') = exp(-\sum_{i =1}^{m} (\phi_{i}(x_{i}-x_{i}^{'})^2) - \sum_{j =1}^{q} ||z_{1,j} - z_{1,j}^{'}||^{2}_2 + ||z_{2,j} - z_{2,j}^{'}||^{2}_2)
\end{equation}

Consequently, the LVGP model parameters $\mu$, $\sigma^2$, and $\phi$, along with the latent variables, $\mathbf{z}$, are estimated through MLE of the  log-likelihood function

\begin{equation}\label{eq:likelihood}
        l(\mu,\sigma,\phi,\mathbf{z}) = -\frac{n}{2} ln(2\pi\sigma^2) - \frac{1}{2}ln|C(\mathbf{\phi},\mathbf{z})| - \frac{1}{2\sigma^2}(\mathbf{y}-\mu\mathbf{1})^{T}C(\mathbf{\phi},\mathbf{z})^{-1}(\mathbf{y}-\mu\mathbf{1})
\end{equation}

where $n$ is the number of available training samples, $C$ is a $n \times n$ correlation matrix with $C_{ij} = c(w_i,w_j)$ for $i,j = 1,2,...,n$, $\mathbf{y} = [y_1,y_2,...,y_n]$ is the observed response vector, $\mathbf{1}$ is a vector of ones with a size of $n \times 1$. Once all the model parameters are estimated, the LVGP model can make predictions, $\hat{y}(\mathbf{w}^*)$, on any given input $\mathbf{w^*}$ with the quantified prediction uncertainty, $\hat{s^2}(\mathbf{w}^*)$:

\begin{equation}\label{eq:y_pred}
        \hat{y}(\mathbf{w}^*) = \mu + \mathbf{c}(\mathbf{w}^*)\mathbf{C}^{-1}(\mathbf{y}-\mu\mathbf{1})
\end{equation}

\begin{equation}\label{eq:sigma_pred}
        \hat{s^2}(\mathbf{w}^*) = \sigma^{2}(\mathbf{c}(\mathbf{w}^*) - \mathbf{c}(\mathbf{w}^*)\mathbf{C}^{-1}\mathbf{c}(\mathbf{w}^*)^{T})
\end{equation}

where $c(\mathbf{w}^*) = [C(\mathbf{w}^*,\mathbf{w}^{(1)}),C(\mathbf{w}^*,\mathbf{w}^{(2)}),...,C(\mathbf{w}^*,\mathbf{w}^{(n)})]$ is the pairwise correlation vector between the new point $\mathbf{w}^*$ and each training point $\mathbf{w}^{(i)}, i = 1,2,...,n$. 

With the introduction of LVGP, the primary contribution of this paper lies in the modeling of the multiple sources of information as a representative qualitative (categorical) input in addition to known quantitative (numerical) inputs through the LVGP model, making the surrogate modeling source-aware and interpretable. Further details of the multi-source data fusion modeling are provided in the subsequent section.

\subsection{Interpretable Multi-Source Data Fusion Through Latent Variable Gaussian Process}

\begin{figure}[t!]
\centering
\includegraphics[width=1\textwidth]{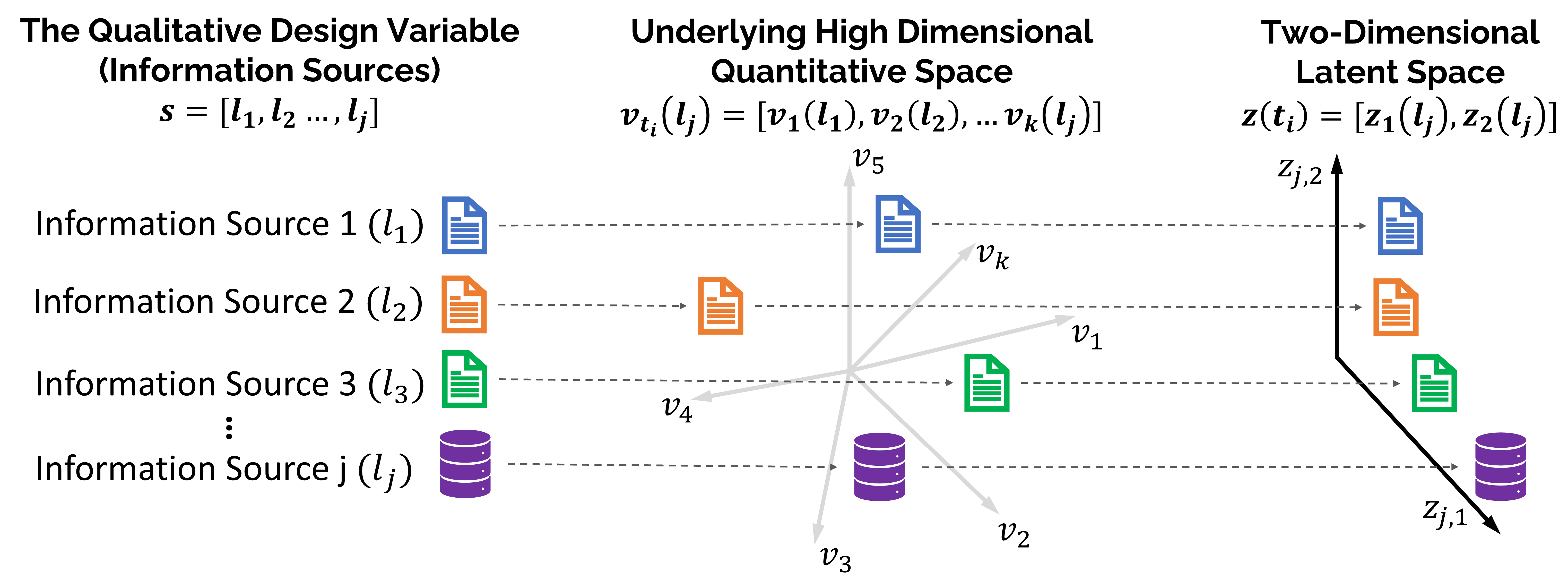}
\caption{Mapping of different information sources (qualitative space) onto the 2-dimensional quantitative latent space}
\label{fig:lvgp_ex}
\end{figure}

\begin{figure}[t]
\centering
\includegraphics[width=0.7\textwidth]{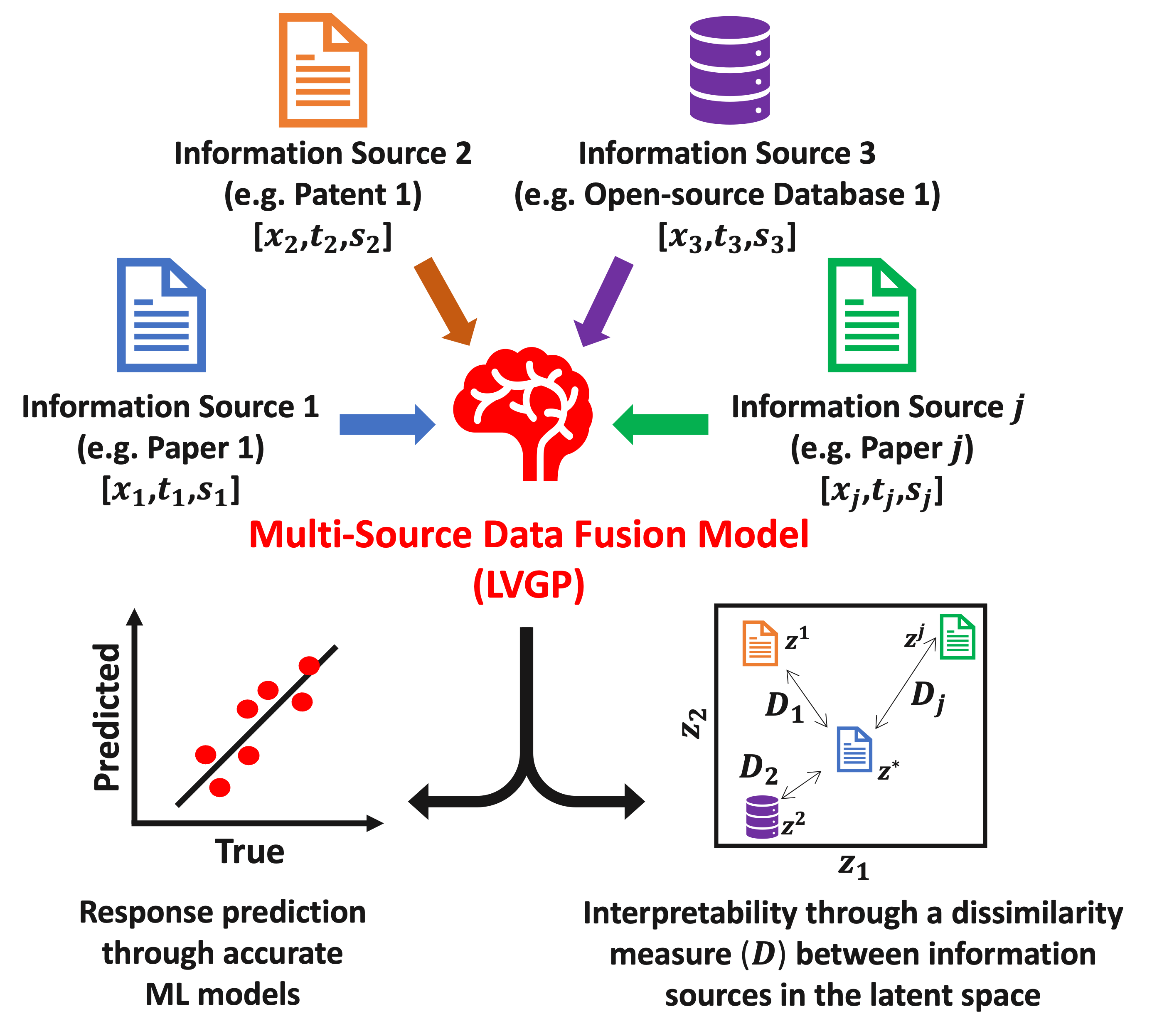}
\caption{Schematic of multi-source data fusion through LVGP}
\label{fig:sch_lvgp}
\end{figure}

For an engineering system of interest, there could be multiple sources of information (patents, papers, open-source repositories) that aim to analyze the system. However, due to the diverse nature of data collection, each information source could contain different underlying procedures and processes that are not reported such as experimental errors, confidential information and so on.  In this work, a multi-source modeling technique with Latent Variable Gaussian Process is proposed to combine all of the information sources into a single source-aware model that considers not only the relationships between information sources but also that both the known and unknown underlying physical parameters associated with each information sources when making predictions. To achieve this, a given engineering system of interest is modeled with a new input space, $[\mathbf{x},\mathbf{t},\mathbf{s}]$, where $\mathbf{x}$ and $\mathbf{t}$ are the known common numerical ($\mathbf{x}$) and categorical ($\mathbf{t}$) system parameters between information sources, and $\mathbf{s}$ is the newly introduced separate categorical variable that represent the information sources ($\mathbf{s}$), along with the output response of interest, $\mathbf{y}$. Here,  $\mathbf{s}$ is treated as a unique separate qualitative variable that differentiates the information sources and integrates them into the multi-source model through the same statistical inference techniques used to estimate the latent variables of the variable $\mathbf{t}$. For this newly introduced variable $\mathbf{s}$, each information source is labelled with the letter $l$. A visual representation of the latent mapping of qualitative variable ($s$), which is denoted as the information sources where each source is represented by a level ranging from $l_1$ to $l_j$  in current case, is shown in Figure \ref{fig:lvgp_ex}. By incorporating information sources as a qualitative variable, the aim is to learn and extract the unknown physical parameters underlying each information source through 2D latent spaces using statistical inference techniques that has been reviewed in Section 2.2.1. It is important here to note that the spatial relations (distances) between the information sources in the latent space can show similarity and differences regarding the influence of the information sources on the system of interest. Similarly, spatial relationships between latent variables can also imply the dimensionality of underlying known and unknown physical parameters related to the information sources. Therefore, for a given information source $j$, it is incorporated into the LVGP model with it's unique known common parameters $x_j$ and $t_j$ and it's unique source identifier $l_j$. Once the LVGP is trained, the model can become aware of not only the known and unknown assumptions and parameters underlying each information source but also the relationships between information sources. The awareness of the sources provided by the LVGP can further lead to fast and accurate predictions. Figure \ref{fig:sch_lvgp} shows the proposed multi-source modeling framework along with information source examples and their unique inputs to the LVGP model. It is important here to note that although common qualitative variables ($\mathbf{t}$) can be integrated into the multi-source model with LVGP, case studies with common quantitative variables ($\mathbf{x}$) are only demonstrated, leading to an input space with only $\mathbf{x}$ and $\mathbf{s}$ that are fed into the LVGP model.

As previously mentioned, the latent variable approach provides physics-based dimension reduction that could potentially explain the "cause-effect" relationship between the inputs (the information sources) and outputs (response). Therefore, the proposed model can give us physical insights regarding the information sources by analyzing the spatial locations of the variables (sources) in the latent space. Here, the locations of the latent variables could suggest that the variables that are closer in the latent space can have a similar impact on the response. Moreover, for the scenarios where no prior knowledge exists regarding information sources, the LVGP model can help us extract and reveal the relationships between the sources. To understand the underlying similarities and differences between information sources, a dissimilarity metric ($D$) that quantifies the dissimilarities between the sources through a distance metric is introduced. This dissimilarity metric is measured through the Euclidean distance of the data source $(z^j)$ from the reference data source $(z^*)$ in the latent space. The metric is given in Equation \ref{eq:D_1}. For a given LVGP model, the latent variables are constrained to be between $z = [-3,3]$ as this was the typical practice in previous applications of LVGP and the reference data source will be placed at $z^* = [0,0]$ in the latent space. Therefore, the most dissimilar information source from the reference source is going to be placed at [$\pm 3,\pm 3$], making $z^{max} =  [\pm3,\pm3]$. Based on this, the maximum distance between the reference and given information source can be $||z^{max}-z^*|| =  3\sqrt{2}$. Thus, a normalized unit of dissimilarity ($D$) can be calculated through Equation \ref{eq:D_2}. With this quantified metric, the similarities and differences between information sources can be understood, which could be further used in downstream tasks such as anomaly detection and fault identification of sources. Finally, although the choice of reference source is determined before the LVGP model training, the reference source choice does not influence the spatial relationships obtained from the latent space. As a result, independent of reference source selection, the same spatial relationship between the sources, hence same $D$ values between sources will be obtained for any reference source selection. Thus, the reference source selection is user, goal, and application specific.

\begin{equation}\label{eq:D_1}
D(z^j) =  \frac{||z^j-z^*||}{||z^{max}-z^*||} 
\end{equation}

\begin{equation}\label{eq:D_2}
D(z^j) =  \frac{||z^j-z^*||}{3\sqrt{2}}
\end{equation}

\section{Representative Mathematical Problems}
Before the proposed methodology is applied into material science case studies, it is first implemented on two representative mathematical problems to provide a more in-depth analysis and understanding. It is demonstrated that when different information sources are identified and specified to the ML models, more accurate models and response surfaces can be obtained, which could be significantly beneficial for model prediction and design optimization purposes.

\subsection{Parabola Problem}

\begin{figure}[t!]
\centering
\includegraphics[width=1\textwidth]{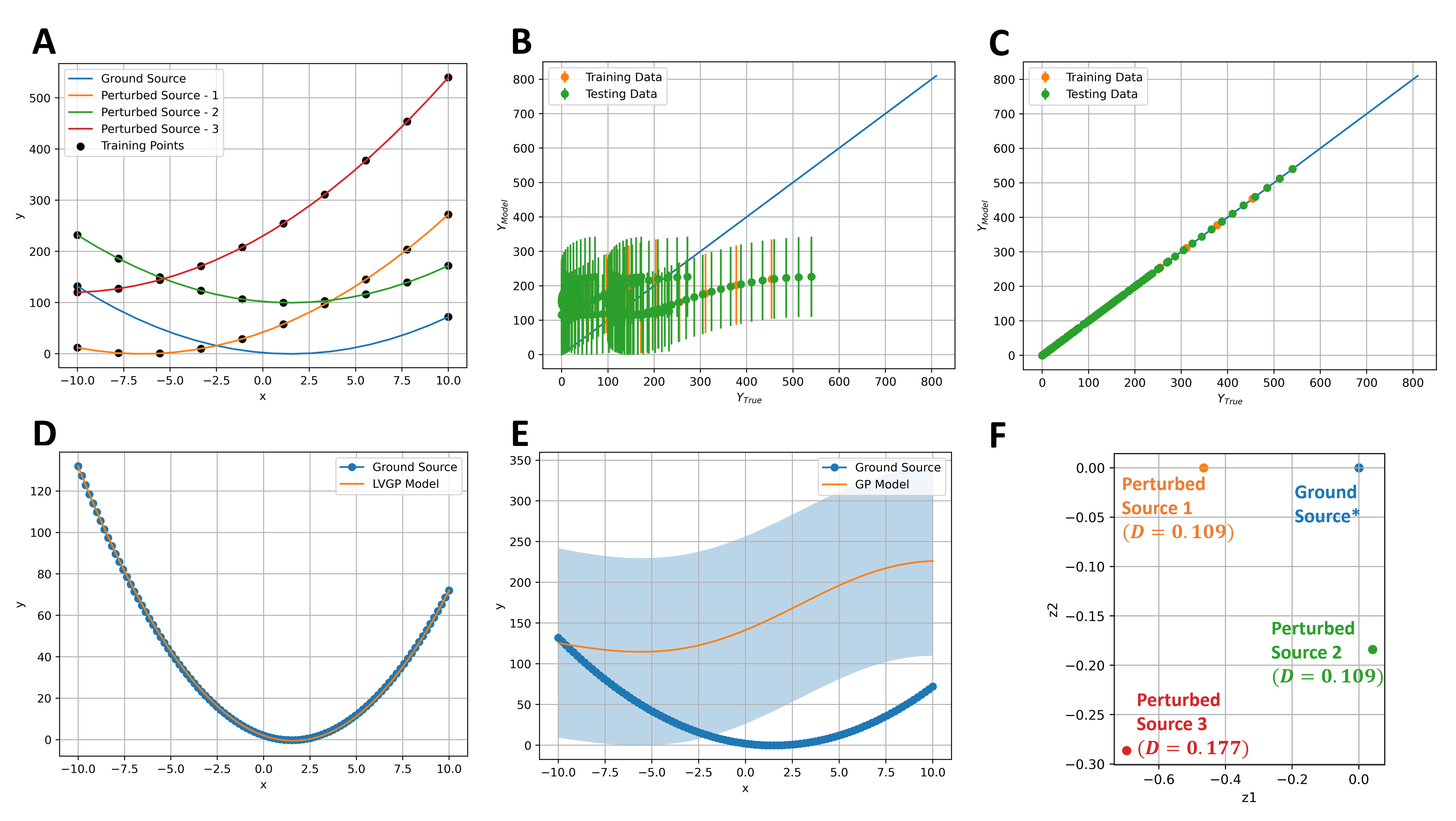}
\caption{(a) Overview of Parabola Example. (b-c) The parity plot between the GP and LVGP models. (d-e) Prediction of Parabola on the Ground source GP model and LVGP model. (f) Latent Space of Parabola Example from LVGP. Ground Source is denoted as the reference source (*) for dissimilarity metric, $D$, calculations}
\label{fig:parabola_summary}
\end{figure}

In the representative parabola problem, four (4) different mathematical equations are used as different information sources. All of the four different information sources are created using 

\begin{equation}
\label{eq:parabola}
    y = (x + x_{shift} - a)\times(x + x_{shift} - b) + y_{shift}
\end{equation} 

where different parameters, $a, b ,x_{shift}$ and, $y_{shift}$ are employed for each of the information sources. For each source, multiple training data points are created between the one dimensional input space, $x \epsilon [-10,10]$. The first information source, which is denoted as the Ground Source, can be considered as the source of data which is most detailed or in-house generated (and replicable) data. Replicating its cost and access, the Ground Source only contains 3 training samples. However, the other three information sources, which can be considered as incomplete or incomprehensive and thus relatively cheap, have 10 training samples each for building the machine learning models. The remaining parabolas are referred to as Perturbed Sources. Supplementary Table \ref{tab:parabola data} shows the parameters used for each of the information sources and details regarding the number of available samples for training and testing the ML models. Furthermore, Figure \ref{fig:parabola_summary}a illustrates the parabolas obtained for each of the information sources. The Perturbed Source 2 is a copy of Ground Source with a $y_{shift}$, Perturbed Source 1 is another copy of Ground Source with a $x_{shift}$, whereas Perturbed Source 3 is a copy of Ground Source with shifts in both directions. The parameters of the perturbed sources are selected carefully to display different resemblance levels to the Ground Source, which will be important for the latent space obtained from the LVGP model and the dissimilarity metric ($D$), as seen later in the manuscript. It can be observed from the Figure \ref{fig:parabola_summary}a that the mathematical and visual resemblance to the Ground Source is in the order of Perturbed Source 2, Perturbed Source 1, and Perturbed Source 3. 

Next, Gaussian Process and Latent Variable Gaussian Process (detailed as GP and LVGP in the previous sections) are leveraged to model the response surface of the representative parabola problem ($y$). For the GP model training, only the quantitative (numerical) input variable ($x$) is used to learn the response. In this case, the model is unaware of the sources where the data is generated from. On the other hand, a separate qualitative variable that denotes different information sources is included into the training of the LVGP model, in addition to the quantitative variable ($x$), making the surrogate model aware of the different sources that the data is generated from. The black points marked on Figure \ref{fig:parabola_summary}a displays the quantitative training points used to train the models. A well-known ML model predictive performance metric, Normalized Root Mean Squared Error (NRMSE) is utilized to compare the training and testing performances of these models. Table \ref{tab:results_table_parabola} shows the performance results comparing the two ML models. The low NRMSE value for LVGP clearly shows that incorporating the information source as a separate variable into data fusion ML models yield a better predictive model. Furthermore, the parity plot of the two models (Figure \ref{fig:parabola_summary}b\& c), comparing the model predictions and actual values, clearly demonstrates the advantage of incorporating and differentiating the source information into the modeling. Finally, Figure \ref{fig:parabola_summary}d \& e shows the predictions made solely on the Ground Source, which once again clearly demonstrates the predictive capabilities of LVGP. In the context of engineering and scientific applications, considering cases where conducting experiments are costly or one has only access to a single experimental resource, obtaining an accurate predictive model for that specific experiment can be highly beneficial for advanced design and development. 

\begin{table*}[h]
\small
\centering
\caption{Predictive performance comparison between GP and LVGP for the parabola problem} \label{tab:results_table_parabola}
\renewcommand{\arraystretch}{1.3}
\begin{tabular}{ccccccccc}
\toprule
Model Type & Training NRMSE  & Testing NRMSE \\
\toprule
GP & $0.199$ & $0.207$  \\
\\
LVGP & $5.26\times10^{-5}$  & $11.66\times10^{-4}$ \\
\toprule
\end{tabular}
\end{table*}

The advantages of LVGP are not limited to providing improved prediction accuracy and the capability of providing uncertainty quantification in the similar way as other GP models. Figure \ref{fig:parabola_summary}f shows the benefit of using LVGP for interpreting the differences between the information sources. The distances between the latent variables of Perturbed Sources to the Ground Source directly relate to the previously mentioned resemblance level to the Ground Source. This resemblance is further quantified with the previously introduced dissimilarity metric $D$. The $D$ values shown in Figure \ref{fig:parabola_summary}f shows the dissimilarity metric for each information source with respect to the Ground Source. It can be inferred from the metric that Perturbed Source 2 is the closest while Perturbed Source 4 is the farthest to the Ground Source. This result also matches very well with our known physical knowledge regarding the information sources and confirms the validity and interpretability of both the latent space and the dissimilarity metric. For this illustrative example, although the relationships between the information sources is known beforehand, the latent variables corresponding to the information source not only revealed the relationship between the sources but also confirmed our initial physical knowledge about the differences between the information sources. Next, the approach is demonstrated on another mathematical testing function used in optimization studies to further demonstrate the importance and effectiveness of using LVGP as a multi-source data fusion method.

\subsection{2D Ackley Function}

\begin{figure}[t!]
\centering
\includegraphics[width=1\textwidth]{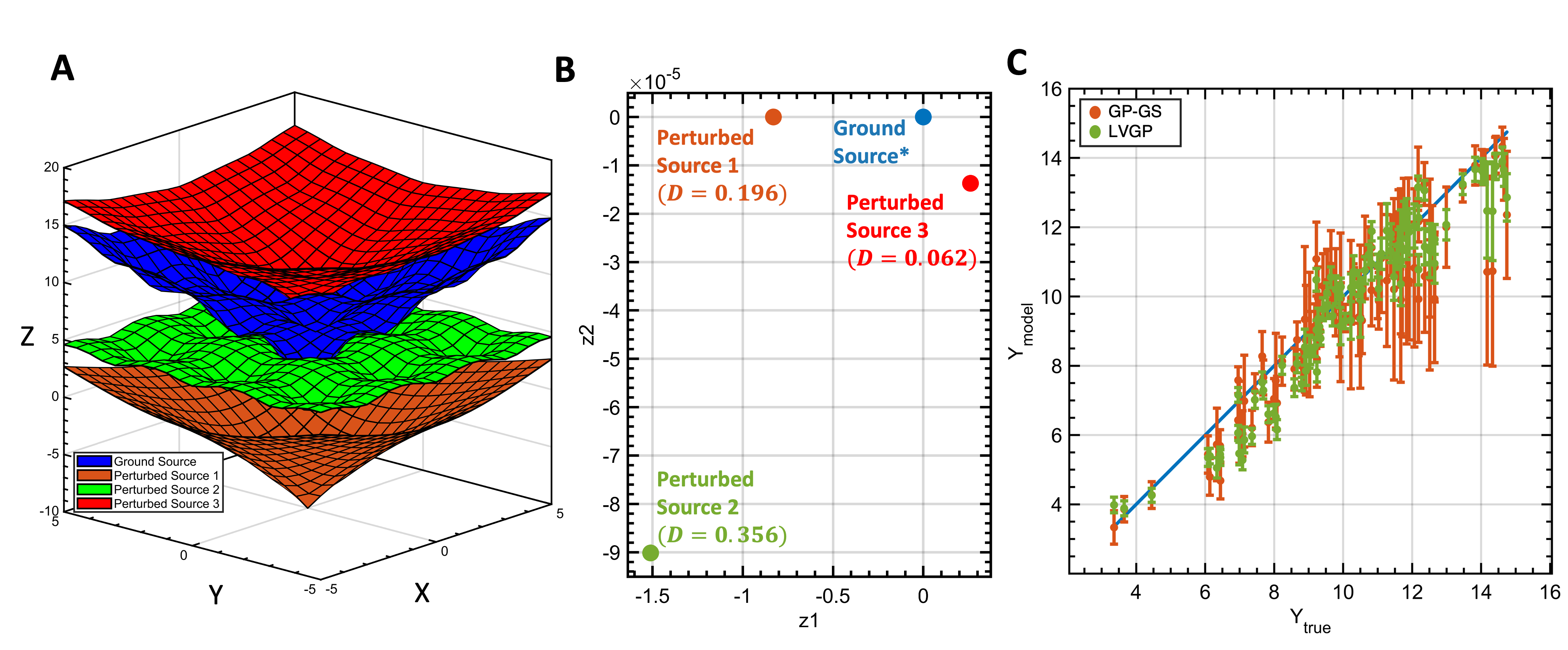}
\caption{(a) Response Surface of the Ground Source and Perturbed Sources for the 2D Ackley's function. (b) The latent variables obtained from the LVGP model. Ground Source is denoted as the reference source (*) for dissimilarity metric, $D$, calculations. (c) The parity plot between the GP-GS and LVGP models tested on the Ground Source data only}
\label{fig:ackley_summary}
\end{figure}

The proposed method is next demonstrated on a well-known mathematical function that is used in optimization studies. Ackley's function is a customizable multi-dimensional function that has been implemented in numerous optimization applications for testing and validation purposes \cite{ackley2012connectionist, ackley2}. Here, 2D Ackley's function, Equation \ref{eq:ackley}, is implemented to further demonstrate the proposed interpretable multi-source data fusion method. In addition to the original 2D function, three (3) other variations of Ackley's function are generated from the original function as different sources of information. Similar to the parabola problem, the original function is named as the Ground Source and the remaining 3 functions are named as Perturbed Sources 1, 2, and 3. The formulations of the functions are found in Supplementary Table \ref{tab:ackley_data} and their surface of the response, $z$, is shown in Figure \ref{fig:ackley_summary}a. 

\begin{equation}
\label{eq:ackley}
    z = -a*exp(-b(\sqrt{\frac{1}{2}(x^2-y^2)})) - exp(\frac{1}{2}(cos(cx)+cos(cy)))+a+exp(1)
\end{equation} 

Identical to the previous case study, two ML models are built (GP and LVGP) to demonstrate the effectiveness of the proposed approach. First, a GP model is built by only using the quantitative variables, $x$ and $y$ to predict the response, $z$. Then, in addition to the two quantitative variables, a qualitative variable is added to differentiate the information sources for the LVGP model. This approach is the exact same approach taken for the previous parabola problem and will be implemented for the remainder of this work. During the training, 5-fold cross validation (CV) technique is implemented and tested on a separate testing set. Both the training and testing sets are created by sampling uniformly from the space $[x,y] \epsilon [-5,5]$. The number of training and testing samples employed to build and validate the models are available in Supplementary Table \ref{tab:ackley_data}. Next, a comparison study between the predictive performance of both models using the NMRSE metric is carried out. Table \ref{tab:results_table_ackley} (Model Types: GP and LVGP) shows the performance metrics for both models. From the results, it can be clearly observed once again that incorporating different sources as a separate variable significantly improves the predictive performance of the model. The model was also able to capture the relationships between the response surfaces and provide interpretability through the latent variables clearly. The latent space obtained from the LVGP, Figure \ref{fig:ackley_summary}b, matches very well with the response surfaces, Figure \ref{fig:ackley_summary}a. Furthermore, the dissimilarity metric calculated using the Ground Source as the reference source, $D$, agree with the response surfaces and what is known regarding the similarity between the functions, demonstrating the efficacy of the method and the metric (Figure \ref{fig:ackley_summary}b).

A possible follow-up question one might ask as a response to the demonstrated results could be to build separate GP models to learn the information sources and predict their responses individually to achieve better accuracies. To answer this question, a separate GP model is built only using the Ground Source and compared with the previously built LVGP on a separate testing set. For ease of notation, this single-source model is denoted as GP-GS. The trained GP-GS model only contained 20 samples from the Ground Source whereas the LVGP model contained all 170 samples from the available sources. The same testing set from the Ground Source is used and the prediction performances are compared. Table \ref{tab:results_table_ackley} (Model Types: LVGP and GP-GS) shows the prediction performances between the GP-GS and LVGP models, along with the parity plot in Figure \ref{fig:ackley_summary}c. In addition to achieving lower NRMSE values for accurate modeling, the parity plot demonstrates that the prediction uncertainty provided by the LVGP model is much less compared to the GP model. This result reveals an interesting phenomenon. Compared to the model that is only trained on the original single-source of information, gathering further information from other sources and explicitly denoting information sources into the data fusion modeling results in a more confident predictive model that provides lower prediction uncertainty on the original single-source of information.

\begin{table*}[h]
\small
\centering
\caption{Predictive performance comparison between GP, LVGP, and GP-GS models on the 2D Ackley function}
\label{tab:results_table_ackley}
\renewcommand{\arraystretch}{1.3}
\begin{tabular}{*{7}{>{\centering\arraybackslash}p{1.85cm}}}
\toprule
Model \qquad Type & Mean Training NRMSE  & Mean CV NRMSE  & Testing NRMSE \\
\toprule
GP & $0.253$  & $0.289$ & $0.254$\\
\\
LVGP & $0.013$  & $0.029$  & $0.024$\\
\\
GP-GS & --- & --- & $0.089$ \\
\toprule
\end{tabular}
\end{table*}

The conclusions from this study can be significant for many reasons and can be applied to many engineering and scientific applications, specifically for materials science. This is because many researchers only have access to or control over their own experiments/simulations and rely on building a surrogate model using their own data to guide their next property evaluations. On the other hand, due to the high costs of property evaluations, data curation for better predictive surrogate models could be expensive and limited to only a few samples. However, by incorporating more data from different available sources (open-sources, papers, patents) and explicitly differentiating each data source as an input to their models, researchers have the potential to obtain better and confident predictive model, which in return has the potential to lead to novel material candidate discovery at a faster rate.  In the following two sections, Section 4 and Section 5, the advantages of using multi-source data fusion through LVGP on two materials design cases are demonstrated.

\section{Modeling Thermal Aging Behavior of FeCrAl Alloys}
\subsection{Overview}

\begin{figure}[t!]
\centering
\includegraphics[width=0.6\textwidth]{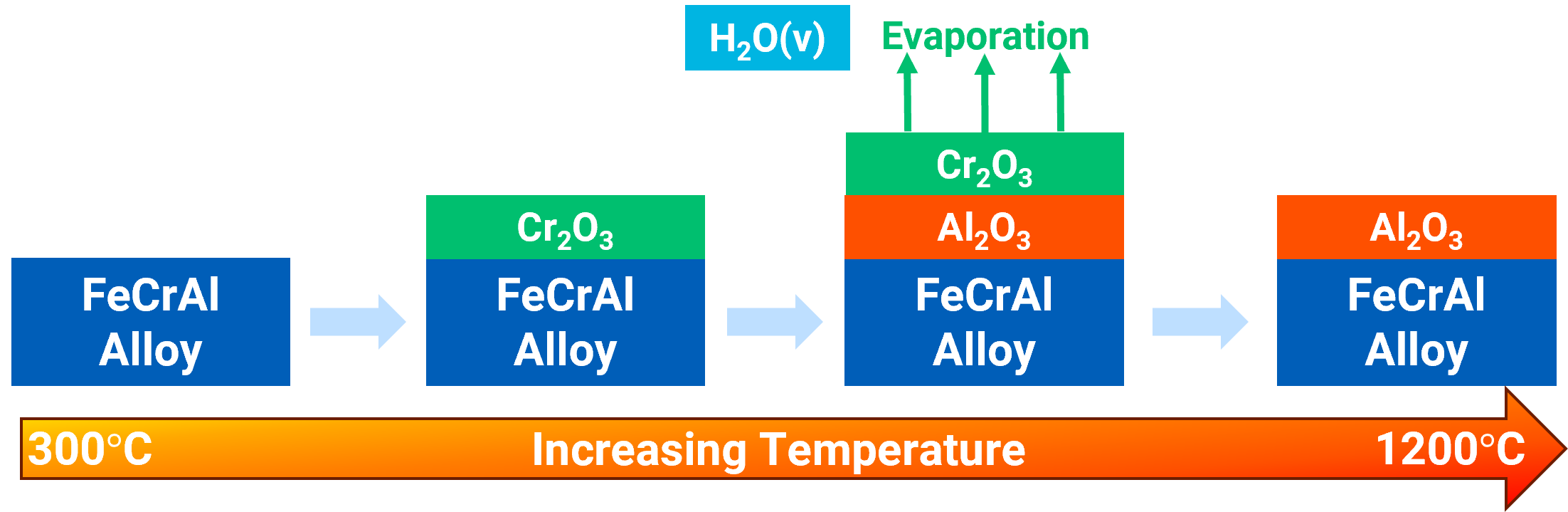}
\caption{Schematic of the oxidation behavior of the FeCrAl alloy}
\label{fig:fecral_mech}
\end{figure}

FeCrAl (Iron-Chromium-Aluminum) alloys have been garnering a lot of attention for their superior hydro-thermal corrosion resistance at normal operating conditions ($\sim$300$^{\circ}$C) and high temperature ($\sim$1000 $^{\circ}$C) steam oxidation resistance \cite{chikhalikar2022effect,kobayashi2010mapping,field2018precipitation,capdevila2008phase,capdevila2008aluminum,li2013effect,han2016effect,hoffman2021effects}. The key application driving the exploration and development of the alloy system is for the protective cladding for nuclear fuels. Their uniqueness lies in their ability of providing oxidation resistance across a wide range of operating temperatures, usually ranging from 300$^{\circ}$C to 1200$^{\circ}$C. A schematic of the oxidation phenomenon is shown in Figure \ref{fig:fecral_mech}. At low operating temperatures (at around 400$^{\circ}$C), Chromium reacts with environmental oxygen and forms a protective oxide layer serving as a natural barrier protecting the alloy from environment. As the operating temperature increases (to around 1200$^{\circ}$C), the Chromium oxide (\ch{Cr2O3}) layer decomposes. The decomposition is subsequently and simultaneously followed by the formation of an Aluminum oxide (\ch{Al2O3}) layer. The Aluminum oxide layer serves as the protective natural barrier against the environment. The whole phenomenon of oxide formation in pure FeCrAl alloys are primarily driven by Chromium content, Aluminum content in the alloy and operating temperatures.

Despite their superior oxidation resistance measured through change in specific mass of the alloy coupon, there is one major hindering aspect of FeCrAl alloys that warrants further study. FeCrAl alloys undergo aging embrittlement which makes them vulnerable to mechanical perturbation and failure. The embrittlement is majorly caused by the Cr-rich $\alpha^{'}$-phase precipitation in the Fe-rich matrix. The formation of such $\alpha^{'}$ phases increases the overall hardness of the alloy and makes them brittle. 

The thermal age hardening phenomenon is driven by the composition of the FeCrAl alloy, temperature, and duration of the thermal aging process. The hardness change measured through Vicker's hardness of the alloy at different stages of thermal aging process is studied and used as an indication of the embrittlement. With the nuance of thermal aging phenomenon in FeCrAl alloys, material designers have the task of optimizing the alloy composition for maximum oxidation resistance and minimum embrittlement. The levers that material designers have at their disposal is the composition of the FeCrAl alloys including dopants that can suppress the formation of $\alpha^{'}$ phases.

To demonstrate the nuances, impact, and need of multi-source data fusion, thermal aging behavior of FeCrAl alloys is used as the first material science case study. In this work, all sources of data are experimental. Recently, there have been several efforts towards leveraging AI$\backslash$ML techniques for modeling and optimizing both the oxidation and thermal aging behavior of FeCrAl alloys \cite{ravi2023elucidating,roy2023understanding,roy2023data,roy2023optimizing}. However, there are still restrictions about the quality of the data from multiple sources and the unaccounted effects of underlying unknown physical parameters that is present in the data generation process during the experimentation. In this study, the aim is to address these challenges using the proposed multi-source ML model.

\subsection{Data Description}
As with all material discovery efforts towards developing a generalizable multi-source data fusion model, the very first step is to gather data from research literature (papers, patents, etc.) and open-source databases. From an extensive search, seven well-cited and verified external sources of data were identified and collected. Additionally, there was also elaborate in-house testing conducted with GE Research to understand the embrittlement behavior of FeCrAl alloys. The sources of data are summarized through Table \ref{tab:fecral data}. The in-house generated samples are tagged as 'GE Data'. The rest of the collected samples are denoted by the name of the respective first authors of the paper. A total of 380 data points has been collected. 

\begin{table}[h!]
\small
\centering
\caption{Sources of information for FeCrAl alloys} \label{tab:fecral data}
\renewcommand{\arraystretch}{1.3}
\begin{tabular}{p{1.5in}p{1.5in}p{3.0in}}
\toprule
Name of Data Source & Number of Data Samples & Source of Data \\
\toprule
GE Data & $178$ & Ravi, Sandipp Krishnan, et al. "Elucidating precipitation in FeCrAl alloys through explainable AI: A case study." Computational Materials Science 230 (2023): 112440. \cite{ravi2023elucidating} \\
Kobayashi & $33$ &  Kobayashi, Satoru, and Takayuki Takasugi. "Mapping of 475 C embrittlement in ferritic Fe–Cr–Al alloys." Scripta Materialia 63.11 (2010): 1104-1107. \cite{kobayashi2010mapping} \\
Han & $55$ & Han, Wentuo, et al. "Effect of Cr/Al contents on the 475ºC age-hardening in oxide dispersion strengthened ferritic steels." Nuclear Materials and Energy 9 (2016): 610-615. \cite{han2016effect} \\
Kim & $4$ & Kim, Hyunmyung, et al. "400 C aging embrittlement of FeCrAl alloys: Microstructure and fracture behavior." Materials Science and Engineering: A 743 (2019): 159-167. \cite{kim2019400} \\
Ejenstam & $12$ & Ejenstam, Jesper, et al. "Microstructural stability of Fe–Cr–Al alloys at 450–550 C." Journal of Nuclear Materials 457 (2015): 291-297. \cite{ejenstam2015microstructural} \\
Yang & $64$ & Yang, Z., et al. "Aluminum suppression of $\alpha^{'}$ precipitate in model Fe–Cr–Al alloys during long-term aging at 475 C." Materials Science and Engineering: A 772 (2020): 138714. \cite{yang2020aluminum} \\
Capdevila & $16$ & Capdevila, Carlos, et al. "Phase separation in PM 2000™ Fe-base ODS alloy: Experimental study at the atomic level." Materials Science and Engineering: A 490.1-2 (2008): 277-288.  and Capdevila, Carlos, Michael K. Miller, and Jesús Chao. "Phase separation kinetics in a Fe–Cr–Al alloy." Acta materialia 60.12 (2012): 4673-4684. \cite{capdevila2008phase, capdevila2012phase} \\
Pinkas & $18$ & Pinkas, Malki, et al. "Sensitivity of thermo-electric power measurements to $\alpha$ – $\alpha^{'}$ phase separation in Cr-rich oxide dispersion strengthened steels." Journal of Materials Science 50 (2015): 4629-4635. \cite{pinkas2015sensitivity} \\
\toprule
Total & $380$ & \\
\toprule
\end{tabular}
\end{table}

\subsection{Model Development}
To contrast and compare multi-source modeling capabilities, GP and LVGP models are trained on the collected data. The inputs for the GP model are the composition of the alloy constituents  (Iron, Aluminum, Chromium, Molybdenum, Zirconium, Manganese, Silicon, Tungsten, Yttrium and Titanium) defined as weight percentages (wt\%), temperature, and duration of the thermal aging process, totaling up to twelve (12) unique inputs. The output that is used to quantify the embrittlement behavior is represented through the measured hardness change (HV). On the other side, the inputs for the LVGP model contain an additional qualitative input that parameterizes each source of data as a distinct categorical level (option) along with all the other 12 numerical inputs used for the GP model. The output for the LVGP is identical to the one used for the GP model.

\subsection{Results and Discussion}
Within this study, (1) the generalizability of the models, and (2) the performances of the models with respect to the experimental in-house information source that can be controlled is investigated. The main goal of (1) is to observe the usefulness of the models when one has control to multiple sources of information to perform further experimentations, whereas the main goal of (2) is to observe which of the models is more reliable when one has further control over a source of information. Out of all 380 available samples, 344 samples are used to observe the generalizability of models through cross validation (CV) and set aside 36 GE Data points as a separate testing set to demonstrate the prediction capabilities of the models with respect to the information source that can be control and perform further experiments on. For the remaining 344 samples, 5-fold cross validation (around 275 training and 69 validation samples per fold) is employed to evaluate the generalizability of both models. For CV tasks in LVGP, stratified sampling is used to make sure datasets used for training and testing represent the overall distribution of the information sources. Then, both models are trained using the 344 samples and tested on the separate 36 GE testing data. The training, validation and testing samples were held identical for both models. The results of this study are shown in Section \ref{fecral:1}. In addition to comparing two ML models on the multi-source data, LVGP is further compared with a separate GP model trained only on GE Data. For ease of notation, this GP model is denoted as GP-GE. GP-GE was only trained on GE Data samples available in the remaining 344 sample set (142 samples), while the previously trained LVGP model on the 344 sample set is used for comparison. Both LVGP and GP-GE are tested on the same separate 36 GE Data samples. The results of this study are shown in Section \ref{fecral:2}. 

Furthermore, the LVGP model is investigated to extract physical insights and to obtain interpretability regarding the sources of information and the system of interest through the latent variables. The results from this examination are highlighted in Section \ref{fecral:3}. Finally, based on insights obtained from the latent space, a targeted source selection through data filtering mechanism is implemented to improve the predictive capabilities of multi-source data fusion in Section \ref{fecral:4}.

\subsubsection{Multi-Source Data Fusion Comparison: LVGP and GP} \label{fecral:1}

The predictive capability comparison between multi-source data fusion through GP and LVGP models are summarized in Table \ref{tab:results_table_fecral} (Model Types: GP and LVGP) and Panels A and B of Figure \ref{fig:fecral_45_plot}. From the summarized prediction metric, NRMSE (Table \ref{tab:results_table_fecral}), although GP model performs slightly better in training instances, we observe a clear improvement in the cross validation (CV) metric using LVGP, which contains the samples that the model has not seen during training. The parity plots from the CV study are shown in Supplementary Figure \ref{fig:fecral_cv}. The CV results demonstrate that incorporating information sources as a separate variable through LVGP can lead to a more comprehensive and generalizable prediction model. Looking further at the testing set which contains separate 36 GE Data samples, LVGP performs significantly better compared to the GP model, demonstrating that explicit categorization of the sources provides more accurate source-specific predictions. Furthermore, the 45-degree parity plot between the GP and LVGP model across the training and testing data (first row of Figure \ref{fig:fecral_overview}) demonstrate that in addition to delivering accurate predictions, LVGP provides lower prediction uncertainties, which can be inferred from the error bars.

\begin{table*}[h!]
\small
\centering
\caption{Predictive performance comparison between GP, LVGP, GP-GE, and LVGP-Target models on FeCrAl alloys} 
\label{tab:results_table_fecral}
\renewcommand{\arraystretch}{1.3}
\begin{tabular}{*{7}{>{\centering\arraybackslash}p{1.85cm}}}
\toprule
Model \qquad Type & Mean Training NRMSE  & Mean CV NRMSE & Testing NRMSE \\
\toprule
GP & $0.060$  &  $0.105$  & $0.088$ \\
\\
LVGP & $0.064$ & $0.100$ & $0.066$\\
\\
GP-GE & ---  & ---   & $0.057$  \\
\\
LVGP-Target & ---  & ---  & $0.053$ \\
\toprule
\end{tabular}
\end{table*}

\begin{figure}[hbt!]
\centering
\includegraphics[width=1\textwidth]{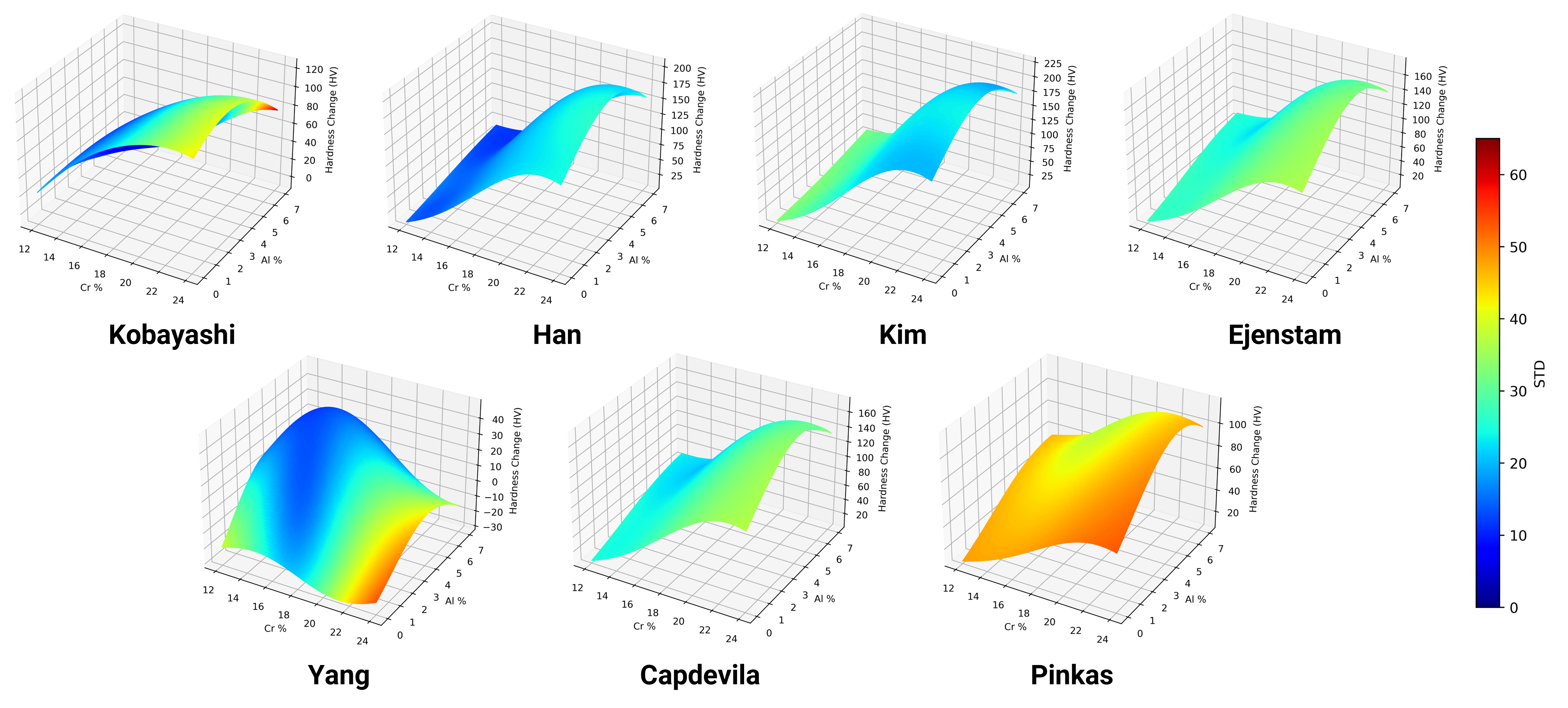}
\caption{Source-specific 3D response surfaces provided by LVGP for the remaining FeCrAl alloy information sources at 12\%<Cr<24\%, 0\%<Al<7\%, 88\% <Fe<69\%, Mo = 0\%, Zr = 0\%, Mn = 0\%, Si = 0\%, W = 0\%, Y = 0\%, Ti = 0\%, Temperature = 475$^{\circ}$C and Duration = 5000 hours, respectively. The colorbar represents the prediction uncertainity as standard deviation. }
\label{fig:fecral_source_specific}
\end{figure}

To strengthen our case of using multi-source modeling through LVGP, response surface plots from both the GP and LVGP models (with GE Data as reference) are also provided through Panels A and B of Figure \ref{fig:fecral_overview}. The second and third row depict the mean and standard deviation of the hardness changes across variation of the Cr (12\%-24\%) and Al (0\%-7\%) and the fourth row depicts the combined mean and standard deviation in a 3D representation. The model validity is assessed by focusing on the mean and variation behavior of the response surface. It is noted that the response surface of the hardness changes are highly undulated with sharp peaks and valleys with the GP model. On the other hand, using the GE source as ground truth for prediction, the response surface is much smoother and more well-behaved with the LVGP model. One can argue that the unsmooth behavior is caused by the varying level of noises coming from different sources in the dataset when fitting all of them to one single GP model. On the other hand, LVGP has the capability of smoothing out the curves from multiple sources after the introduction of the latent variables. From a material science perspective, it is very unlikely that phenomenon driving the hardness changes across the composition are highly transient and acute. This important observation highlights the need of accounting for known and unknown underlying physical parameters when data from numerous sources are combined for data fusion. The samples collected from different papers might have had differences in manufacturing process, thermal aging procedure, and testing technique that are not accounted for in a single source model. Furthermore, the downstream tasks of optimization, sensitivity analysis, uncertainty quantification would also be severely affected by such distinctions and sharp undulations of the surrogate models, thus making the case of employing multi-source modeling through LVGP once again. Finally, LVGP has the capability of providing source-specific predictions, which is not plausible with a single GP model that cannot differentiate between the sources. The response surface predictions provided by LVGP model for the remaining sources of information are provided in Figure \ref{fig:fecral_source_specific}. This property provided by the LVGP can be highly advantageous for the cases when one has control over multiple sources of information.

\subsubsection{Comparison of the Multi-Source LVGP with the Single-Source GP-GE} \label{fecral:2}
\begin{figure}[hbt!]
\centering
\includegraphics[width=1\textwidth]{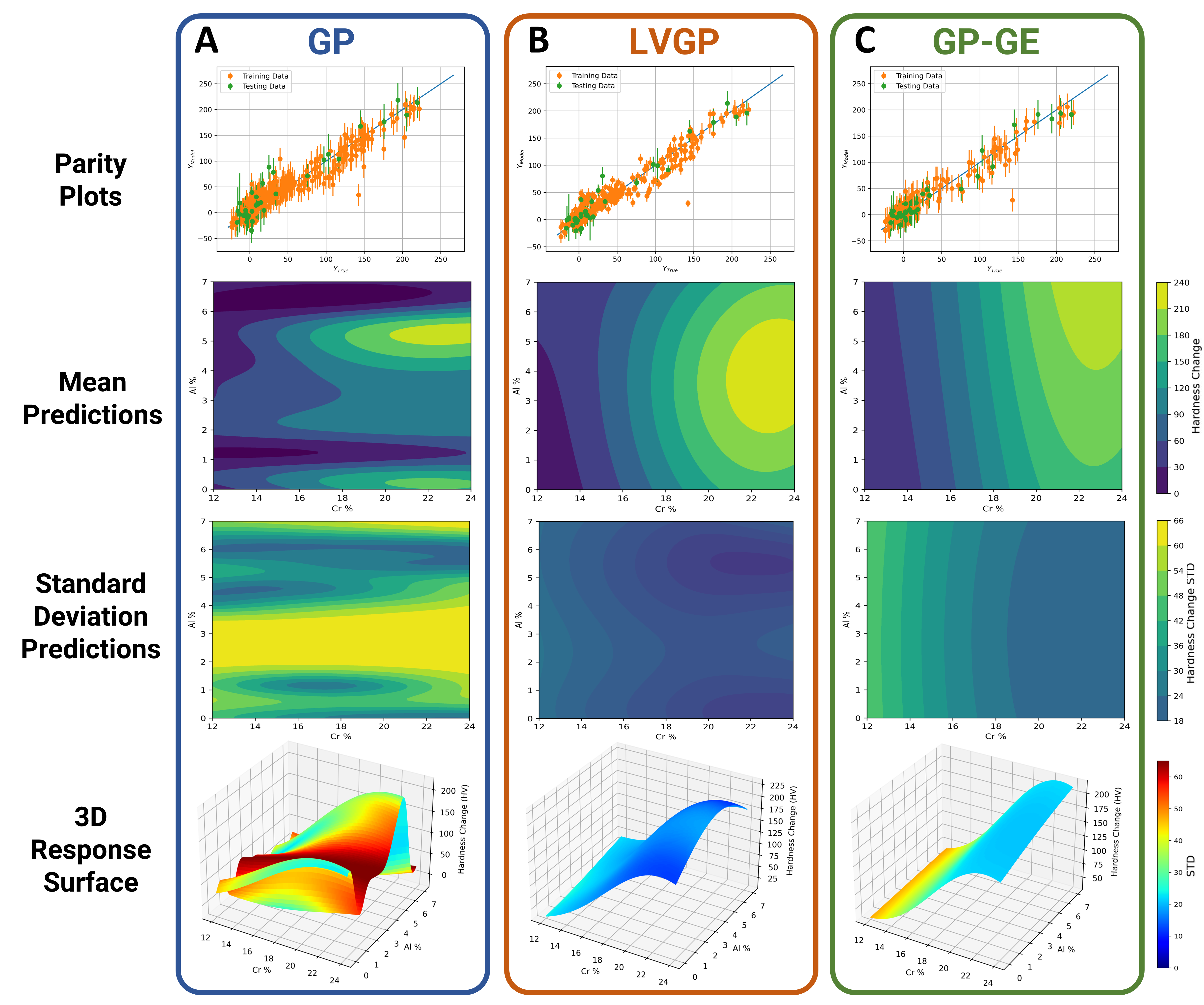}
 \caption{Predictive performance comparison between (a) GP, (b) LVGP, and (c) GP-GE models. The first row shows the parity plots of the models with respect to their respective training and GE Data testing sets. The x and y axis show the true and predicted hardness change (HV), respectively, and error bars demonstrate the prediction uncertainty associated with each prediction. Second, third and fourth row show the Mean, Standard Deviation and 3D Response Surfaces of the models at 12\%<Cr<24\%, 0\%<Al<7\%, 88\% <Fe<69\%, Mo = 0\%, Zr = 0\%, Mn = 0\%, Si = 0\%, W = 0\%, Y = 0\%, Ti = 0\%, Temperature = 475$^{\circ}$C and Duration = 5000 hours, respectively}
\label{fig:fecral_overview}
\end{figure}

A possible argument to achieve more accurate ML models is to build models only using the data source that one has control over, which would be the GE Data in the current study. To demonstrate the advantages of using multi-source data fusion through LVGP over a single-source model for the specified source of interest, the prediction performances of LVGP and the GP model trained on the GE Data only, which was previously denoted as GP-GE, are compared. Table \ref{tab:results_table_fecral} (Model Types: LVGP and GP-GE) and Panels B and C of Figure \ref{fig:fecral_overview} demonstrate the prediction metrics and the response surface obtained from the models. Due to the differences in the data used to train models (multi-source vs single-source), testing accuracy of the models are studied on the common 36 GE Data testing set. The NMRSE prediction metrics from Table \ref{tab:results_table_fecral} demonstrate that incorporating other available sources into the modeling yield very similar predictive performances, with slightly better results towards GP-GE model. However, comparing the 45-degree parity plots between the two models (first row of Figure \ref{fig:fecral_overview}), we observe that the prediction uncertainties provided by the LVGP are much smaller compared to the prediction uncertainties provided by the GP-GE model. The distribution of predicted uncertainties for both models are provided in Supplementary Figure \ref{fig:fecral_std_si}. A possible explanation for this phenomenon could be that the knowledge incorporated from other sources contributes to the learning of the material system and therefore increases the confidence in the model when making predictions. Furthermore, the response surfaces between the two models show high similarity, indicating that the LVGP was able to capture the response surface of the specific source even when other data sources exist. The final very important observation is that the predictions provided by the LVGP model possess larger and lower maximum mean prediction values with lower uncertainties (standard deviation predictions) at different locations compared to both GP and GP-GE models. This observation demonstrates that the LVGP is able can perform extrapolation with much higher confidence by learning from other sources. This result could be significant and advantageous for materials design applications as it can give a multi-step head start to the design optimization task through more confident and accurate predictions.

For research transparency, it is important to note that without further experimental testing and validation in the highlighted compositional space, it cannot be stated which model is a true and generalizable model. The goal of the current work is to illustrate the existence of data discrepancies that could harm multi-source modeling and how they can be integrated into modeling using LVGP. However, the current research study demonstrates that using LVGP as a multi-source data fusion model can be advantageous due to providing (1) smoother response surfaces closer to real physical material behavior, and (2) accurate and confident predictions which can be highly beneficial for downstream tasks such as material optimization, sensitivity analysis, and uncertainty quantification.

\subsubsection{Interpretability from the Latent Variables}\label{fecral:3}

\begin{figure}[hbt!]
\centering
\includegraphics[width=.9\textwidth]{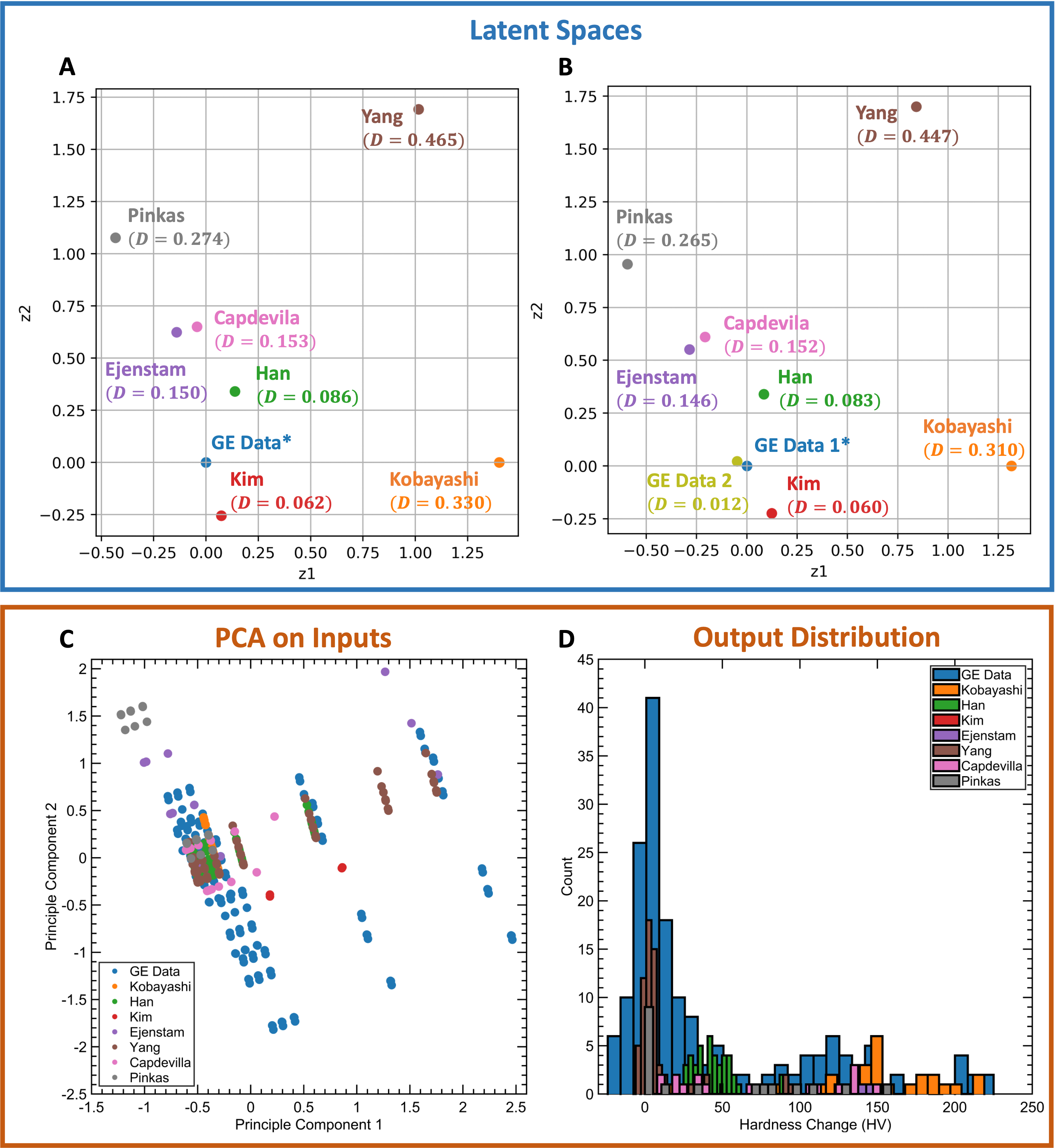}
\caption{(a) The latent Space of FeCrAl sources from LVGP. GE Data is denoted as the reference source (*) for dissimilarity metric, $D$, calculations. (b) The latent Space obtained from the interpretation validation study. GE Data 1 is denoted as the reference source (*). (c) Two principal components of the input space and (d) output distribution of HV with respect to information sources}
\label{fig:fecral_45_plot}
\end{figure}

In addition to providing smoother response surfaces closer to the physical behavior, the LVGP model provides interpretability into different sources through the latent space and the quantified dissimilarity metric, $D$. The obtained latent variables are shown in shown in Figure \ref{fig:fecral_45_plot}a. As previously highlighted, the spatial relations (distances) between the information sources can reveal certain information and provide further interpretability and insights on the sources. For analysis purposes, the GE Data source is used as reference source and compared with rest of the sources as GE Data source can be re-manufactured, optimized, validated, and tested further. It is observed from the latent space that the placement of the sources varies significantly with respect to the reference source. Specifically, the top three sources of data that are dissimilar (higher $D$ value) from the GE Data source are Yang, Kobayashi, and Pinkas, whereas the top three sources of data that are similar (lower $D$ values) to the GE Data source are Han, Kim, and Ejenstam. 

The spatial differences in the latent space can be attributed to differences in the input space, variations in the output space, or underlying fundamental differences in the input-output relationships that cannot be captured from the data. To investigate and interpret the latent spaces, a deeper study on the input and output (property) space is conducted. First, a principal component analysis (PCA) is performed to observe the differences in the input space. Figure \ref{fig:fecral_45_plot}c shows that the two principal components of the GE Data source cover and extend beyond the principal components of the other sources, indicating that the neither the closeness or the remoteness of the sources to the GE Data source in the latent space can be attributed to the differences in the input space. Looking further into the output distributions (Figure \ref{fig:fecral_45_plot}d), it is seen that the GE Data once again covers and extends beyond the properties of other available sources, demonstrating that the differences in the latent space are not due to the variabilities in the output space. As a result, the most impactful observation made from the latent space of data sources is that there are some fundamental differences between the data sources that are not explicitly known. As mentioned earlier, these possible differences for the current case can be due to differences in the manufacturing process of the alloy, testing methods, measuring procedure, underlying phenomenon, and more. Thus, it can be concluded that important underlying differences do exist between the information sources and LVGP is not only able to capture these variabilities but also learns the relationships between the sources.

A further validation study is done by splitting the GE Data into two distinct sources, assigning them different categorical levels as GE Data 1 and GE Data 2, and retraining the LVGP model to observe the latent space again. The aim of this validation study is to observe the new latent space when the same underlying experimental procedures are implemented for two resources along with other sources that have different unknown underlying procedures. The new constructed latent space is shown in Figure \ref{fig:fecral_45_plot}b. A significant observation from the validation study is that the latent variables of GE Data 1 and GE Data 2 are located adjacent to each other. This result was expected since the same experimental procedures are implemented for both sources. However, the results also once again demonstrated and validated that the LVGP model is able to capture the differences in the known and unknown underlying parameters that influence the material system and learn the relationships between the sources. This remarkable capability provided by the LVGP can be highly important for further downstream tasks including but not limited to model validation, dimension reduction, sensitivity analysis, and more. It is important to highlight that without using the LVGP approach proposed in this paper, it would be difficult to learn and assess the relationships between the data sources.

\subsubsection{Enhancing Modeling Capabilities Through Targeted Source Selection Based On Latent Variables} \label{fecral:4}

It is generally acknowledged in the ML community that the prediction performances of the models improve when more data is available. Thus, there has been great efforts of data collection and curation in the ML community. Although this notion of more the data better the model is generally true, there could be cases where more data can actually hurt the ML models and result in poor prediction results. This phenomenon can be especially true for ML applications in materials science where the fundamentals of the generated data, specifically for experimental data, are usually either unknown to the model builders or not disclosed by the data generators due to many reasons such as errors, confidentiality and so on. Therefore, a major challenge still exists regarding the quality, and applicability of the available data. Thus, during the ML model building process, researchers are more inclined to trust and use the data they either generate themselves or they know the fundamentals of. However, this comes with two challenges. First, generating materials data whether experimental or simulation can be expensive and time consuming, which could result in poor ML models further down the line due to low amounts of available training samples. As a result, researchers also rely on other data sources to build their models, in addition to using the data they generate themselves. However, as mentioned previously,  it is not always plausible to know the quality and the fundamentals of the data generation process, meaning that there could be notable differences and discrepancies in the data sources which could significantly influence the performance of the ML models.

So far, it has been demonstrated that the LVGP modeling brings numerous advantages to multi-source data fusion compared to some of the alternative approaches. One of the key advantages of the LVGP comes from the physically interpretable latent spaces that can possibly explain the relationships and differences between different sources of information. As a result, it is proposed that the latent space provided by the LVGP model can act as both a validation and an anomaly detection mechanism to remove the data samples from the sources that could potentially harm the ML model that will be used for further optimization tasks on the source of information that one has control over. For the data sources available in the FeCrAl alloy study, the authors generated and only have further control over the GE Data source. Therefore, the further material design optimization tasks can only take place using the experimental procedures available at GE Research. However, to improve the quality of the ML modelling, numerous data sources have also been collected and used for LVGP model building. Looking at the latent spaces from LVGP model in Figure \ref{fig:fecral_45_plot}a, it is observed that some data sources, Yang, Kobayashi, and Pinkas to be specific, are located at a much further point (higher D values) compared to the other sources when GE Data is selected as the reference point. The stark differences are also observed in the source-specific response surfaces learned and provided by the LVGP (Figure \ref{fig:fecral_source_specific}). As discussed before,  this result can indicate that the experimental procedures taken place by those sources can be different than the ones that are implemented for GE Data. However, this does not necessarily state that there are fundamental errors or data quality issues regarding the three data sources but could potentially indicate that the sources closer to the GE Data can contribute to model accuracy whereas others can harm the ML model for predictions on GE Data. Therefore, to investigate the influence of targeted source selection on the separate GE Data testing set, the three sources that are further from the GE Data (Yang, Kobayashi, and Pinkas), are removed from the training sets and the LVGP model is re-trained with the remaining 229 samples and evaluated on the same GE Data testing set. For ease of notation, the new LVGP model is denoted as LVGP-Target. Table \ref{tab:results_table_fecral} shows the comparison between the prediction accuracy made by the three models, LVGP, GP-GE, and LVGP-Target on the separate 36 GE Data testing set. Although both LVGP and GP-GE have already demonstrated remarkable accuracy results on the GE Data, LVGP-Target provides notable improvement in accuracy, specifically reducing NRMSE by almost 20\% and 7\% in comparison to LVGP and GP-GE models, respectively. The parity plot of the predictions made by the LVGP-Target model are shown in Supplementary Figure \ref{fig:fecral_gptarget}. Considering that the evaluation of properties through experimentation can be relatively expensive, the notable increase in accuracy provided by the targeted source selection through LVGP can provide significant cost and time savings for further downstream tasks. Overall, in addition to providing accurate data fusion modeling, LVGP can be employed as a validation and data filtering mechanism to improve multi-source modeling.

\section{Modeling Magnetic Behavior of SmCoFe Alloys}
\subsection{Overview}
Electric and hybrid electric transportation has been garnering a lot of attention in recent times. One of the key focuses has been developing more denser and powerful motors to provide the propulsion capacity needed. To increase the power density and capacity of the motor, the housed permanent magnet plays an essential role. With the direct increase in magnetic characteristics of fundamental magnetic material, high operational envelope of motors can be achieved. An overview of performance of magnetic alloy systems is given through \cite{cui2018current}. The work identifies alloys that have higher magnetic properties measured through saturation magnetization and that have higher operating temperatures measured through Curie temperature. From the array of magnetic materials, it was observed that SmCo (Samarium-Cobalt) alloy, also referred as SmCoFe, exhibit relatively high stability at high temperatures which is essential to withstand the hot and harsh environments of aircraft engines. Hence the research community has favored the exploration of SmCoFe systems for aerospace application.

There have been numerous efforts that seek to identify the composition and the unique dopants that can improve the energy product envelope of SmCoFe systems catering to hybrid electric flight applications. Similar to our previously presented FeCrAl case study, with the advent of AI$\backslash$ML, data-driven approaches have been extensively explored and used for optimization and modeling. To demonstrate another key nuance and the differentiation of multi-source modeling, magnetic behavior of SmCoFe alloys is used as the second material science case study to address the restrictions about the quality of data from multiple sources and unaccounted effects of known and unknown underlying physical parameters.

\subsection{Data Description}
As with the previous materials case study, the first step is to collect data from open research literature such as papers, patents, and other sources of information that are restricted to SmCoFe alloy system. From an extensive search, three primary sources of information were identified. Unfortunately, the in-house data testing for SmCoFe systems is not readily available but currently on-going. Table \ref{tab:smcofe data} summarizes the sources of data. A total of 71 data points were collected. One distinction from the previous case study is that the data from the one of the patents is further differentiated into two information sources (Patent T2 and Patent T4). The differentiation was necessitated by additional changes to the post-processing parameters that were implemented. 

\begin{table*}[h]
\small
\centering
\caption{Sources of information for SmCoFe alloys}
\label{tab:smcofe data}
\renewcommand{\arraystretch}{1.3}
\begin{tabular}{p{1.5in}p{1.5in}p{3.0in}}
\toprule
Name of Data Source & Number of Data Samples & Source of Data \\
\toprule
Patent T2 & $20$ & Table 2 from Fujiwara, Teruhiko, Hiroaki Machida, and Hideyuki Yoshikawa. "Rare earth-cobalt permanent magnet." U.S. Patent Application No. 14/643,875. \cite{US20150262740A1} \\
Patent T4 & $21$ & Table 4 from Fujiwara, Teruhiko, Hiroaki Machida, and Hideyuki Yoshikawa. "Rare earth-cobalt permanent magnet." U.S. Patent Application No. 14/643,875. \cite{US20150262740A1} \\
Prop of SmCeCoFeCu & $30$ &  Senno, Harufumi, and Yoshio, Tawara. "Permanent-magnet properties of Sm-Ce-Co-Fe-Cu alloys with compositions between 1-5 and 2-17." IEEE Transactions on Magnetics 10.2 (1974): 313-317. \cite{senno1974permanent} \\
\toprule
Total & $71$ & \\
\toprule
\end{tabular}
\end{table*}

\subsection{Model Development}
Similar to the FeCrAl model development approach, GP and LVGP models are trained on the collected data. The common inputs for the GP and LVGP models are the composition of the alloy constituents (Samarium, Zirconium, Copper, Iron, Cerium, Cobalt) defined as the weight percentages (wt\%) and the testing temperature of the alloy, leading to seven (7) unique input parameters. The property of interest (output) for this case study is the energy product (EP) which is an experimentally measured metric that reflects the energy density and power capacity of the magnet. As with the previous case study, the inputs for the LVGP contain an additional qualitative input characterizing each source of data as distinct categorical level along all the other 7 inputs used for the GP model. The outputs for both the models remain the same. 

\subsection{Results and Discussion}
Two separate studies are conducted as previously done with FeCrAl study. Within the first study, generalizability, and the prediction capability of the models through one source that could be controlled are investigate. In this case, although there are no in-house experiments that could be controlled, Prop of SmCeCoFeCu is selected as in-house experiments (reference information source) as it is the source with the highest number of available samples.  First, generalizability of GP and LVGP is compared through 5-fold cross validation using the 65 samples out of 71 available samples, and testing the models over a 6 sample separate testing set that are selected from the Prop of SmCeCoFeCu source. For CV tasks in LVGP, stratified sampling is used to make sure datasets used for training and testing represent the overall distribution of the information sources. The training dataset and testing dataset remain identical for both models. The results of this study are highlighted in Section \ref{smcofe:1}.

In the second study, a separate single-source GP model is built using only the Prop of SmCeCoFeCu source and compared with the multi-source LVGP model. The performances of the models are compared according to the same 6 separate testing set. For ease of notation, this single source GP model is named as GP-Prop. The results of this study are demonstrated in Section \ref{smcofe:2}.

\subsubsection{Multi-Source Data Fusion Comparison: LVGP and GP} \label{smcofe:1}
\begin{figure}[hbt!]
\centering
\includegraphics[width=1\textwidth]{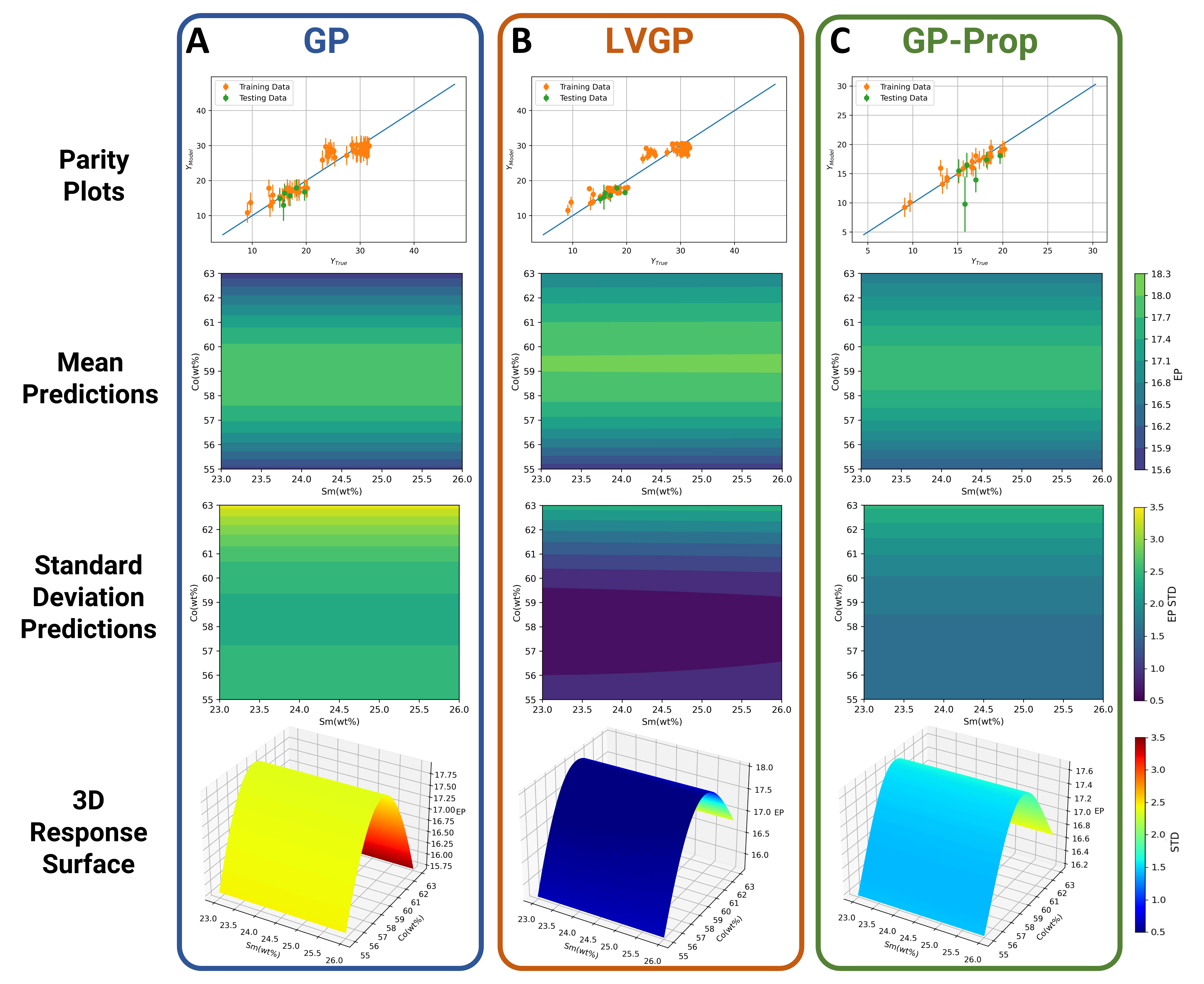}
 \caption{Predictive performance comparison between (a) GP, (b) LVGP, and (c) GP-Prop models.  First row shows the parity plots of the models with respect to their respective training and Prop  SmCeCoFeCu testing sets. The x and y axis show the true and predicted energy potential (EP), respectively, and error bars demonstrate the prediction uncertainty associated with each prediction. Second, third and fourth row show the Mean, Standard Deviation and 3D Response Surfaces of the models at 23\%<Sm<26\%, 55\%<Co<63\%,  1.0\%<Zr<2.0\%, Cu = 8.0\%, Fe = 19.0\%, Hf = 0\%, Ce = 0\% and Temperature = 25$^{\circ}$C, respectively}
\label{fig:smcofe_overview}
\end{figure}

Table \ref{tab:results_table_magmat} summarizes the performance of the GP and LVGP model on the SmCoFe alloy dataset. From the comparison of the NRMSE across the training and cross validation datasets, it can be observed that the multi-source LVGP model shows significant improved performance over the GP model. Therefore, it can be concluded that explicitly including the source of data could enable accurate and generalizable models. Looking at the testing NRMSE metric, it is observed once again that that LVGP model outperforms the GP model, demonstrating it's predictive capabilities. The parity plots from the CV study are also shown in Supplementary Figure \ref{fig:smcofe_cv}. Moreover, from 45-degree parity plots, it is observed that both models make similar predictions on both the testing and training set (Panels A and B of Figure \ref{fig:smcofe_overview}). On the other hand, it can be clearly observed that LVGP provides lower prediction uncertainties as the predictions have lower confidence bounds. Once again, this could be highly advantageous for further downstream materials design optimization studies as the information from other sources can provide a more informed and confident ML model.

\begin{table*}[h!]
\small
\centering
\caption{Predictive performance comparison between GP, LVGP, and GP-Prop models on SmCoFe Alloys}
\label{tab:results_table_magmat}
\renewcommand{\arraystretch}{1.3}
\begin{tabular}{*{7}{>{\centering\arraybackslash}p{1.85cm}}}
\toprule
Model \qquad Type & Mean Training NRMSE & Mean CV NRMSE & Testing NRMSE \\
\toprule
GP & $0.098$  & $0.285$  & $0.078$  \\
\\
LVGP & $0.099$  & $0.155$ & $0.062$ \\
\\
GP-Prop & --- & ---  & $0.258$  \\
\toprule
\end{tabular}
\end{table*}

\begin{figure}[hbt!]
\centering
\includegraphics[width=.5\textwidth]{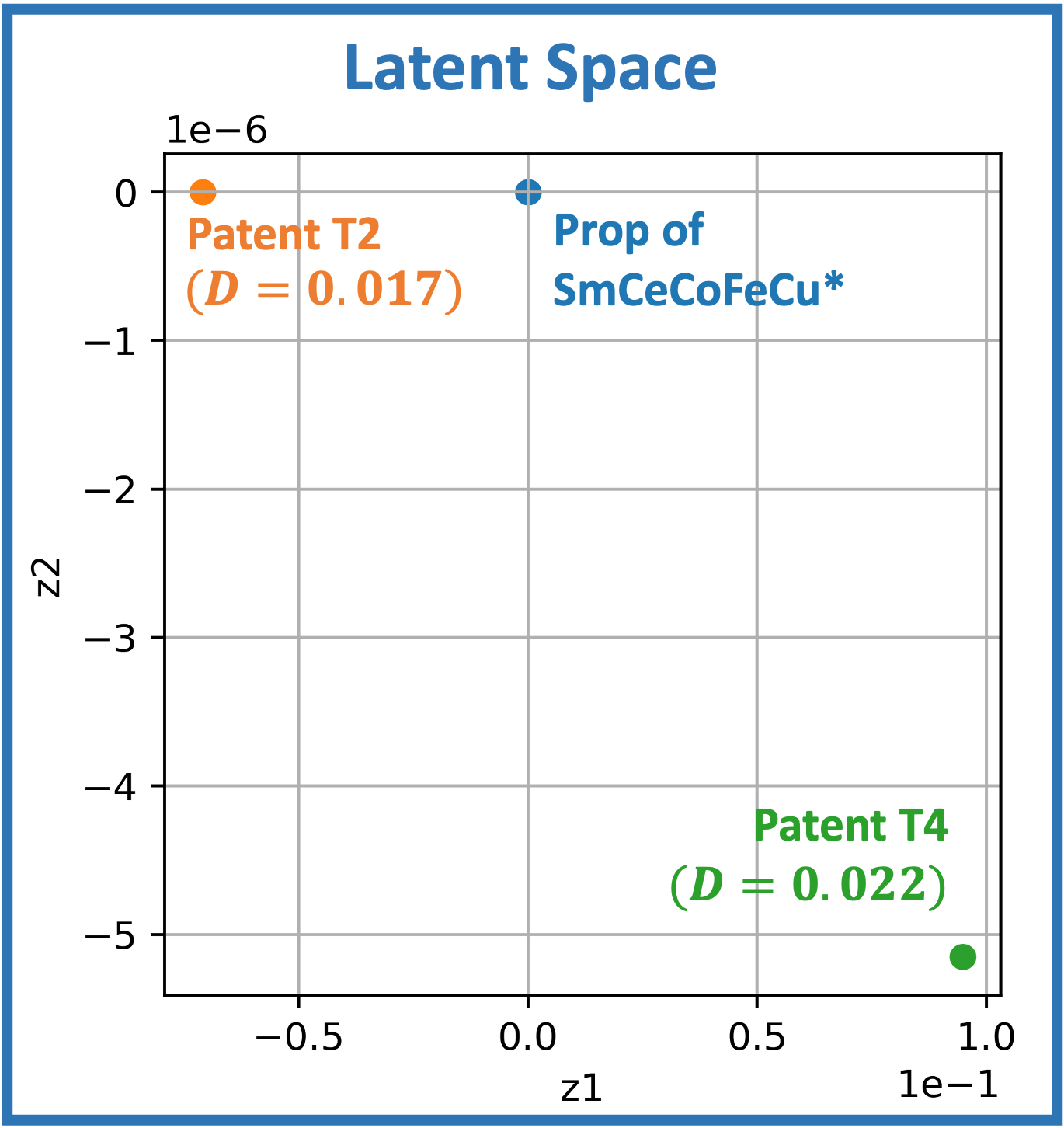}
\caption{The latent space of SmCoFe information sources from the LVGP model. Prop SmCeCoFeCu is denoted as the reference source (*) for dissimilarity metric, $D$, calculations}
\label{fig:smcofe_latent}
\end{figure}

Extending the investigation of the models, the response surface from both the models are plotted across the same input space of the Prop SmCeCoFeCu source through Panels A-B in Figure \ref{fig:smcofe_overview}. The response surfaces (with Prop SmCeCoFeCu as reference due to details and nuances provided in the source) of the models are smooth and well-behaved. The well-behaved nature of the response surface can be attributed to the low dissimilarity measure of each source given through the latent space in Figure \ref{fig:smcofe_latent}. This could indicate that all the sources of information possess high similarities. It is highlighted that since the experiments for Patent T2 and Patent T4 have been provided through the same Patent \cite{US20150262740A1}, it is plausible to assume that the underlying unknown physical experimental parameters are the same for both sources of information and that LVGP was able to identify this explicitly. Furthermore, in addition to providing better predictions, a case is still made for the need of multi-source data fusion through LVGP. Within the highlighted compositional space of the response surface, the location of maximum energy product from GP and LVGP are different. For the GP model, the maximum energy product is around 17.75 MGOe with a prediction uncertainty of around 2.5 MGOe, approximately at Co(wt\%) = 60. For LVGP models centered at Prop SmCeCoFeCu, the maximum energy product is around 18.00 MGOe with an uncertainty of 0.50 MGOe at approximately Co(wt\%) = 62. This nuanced difference is seemingly tied to the source of the data as datasets and models are identical except for inclusion of data source. In addition to higher mean predictions provided by the LVGP compared to the GP model, the predictions also come with much lower uncertainty. Furthermore, for the cases where the designer has access and control over one source of information for further design optimization, the optimum candidates provided by the GP model can be misleading and can misguide the researchers towards wrong optimum designs since there is no way to explicitly differentiate the information sources. Therefore, the reference source selection capability present in the LVGP plays an essential role in plausibly guaranteeing the optimized behavior as it is tied to the most informative source of data (identical procedures can be followed). Additionally, within the compositional space there is a difference of 2\% in Co (wt\%) for the maximum EP property between both models. This shift in the composition of a key rare earth element can have a significant impact in realizing such a superior alloy and downstream cost and supply chain setup.

\subsubsection{Comparison of the Multi-Source LVGP with the Single-Source GP-Prop} \label{smcofe:2}

To overcome the challenge of performing optimization in the unknown information source space, one can argue to build GP models on each information source individually. To demonstrate the effectiveness of multi-source LVGP once again, it is compared against a single-source GP model that is built on the Prop SmCeCoFeCu source only. The single-source GP model, which is previously denoted as GP-Prop, uses the entire available training data from the Prop SmCeCoFeCu source (24 samples) for model building, and later tested on the separate 6 samples obtained from the Prop SmCeCoFeCu once again. The prediction metrics and the parity plots on testing data are shown in Table \ref{tab:results_table_magmat} (Model Types: LVGP and GP-Prop) and Panels B and C of Figure \ref{fig:smcofe_overview}, respectively. From the NRMSE metric, parity plot, and predicted uncertainty distributions (Supplementary Figure \ref{fig:smcofe_std_si}), the clear advantage of using LVGP is observed as it provides much better predictions with lower prediction uncertainties. Furthermore, looking at the response surfaces of both models, Panels B and C of Figure \ref{fig:smcofe_overview}, a tone stark difference is seen in the uncertainty. For LVGP, the uncertainty for majority of the surface is at 0.5 MGOe. For the GP-Prop, the uncertainty for the majority of the surface is at 1.5 MGOe, three times of what is observed for LVGP. The high reduction in uncertainty of LVGP model comes from adding in other data sources. This phenomenon could be a crucial differentiator when designing next generation materials with restrictions on data-availability driven by manufacturing and experimental testing capabilities.

For full transparency purposes, it is once again highlighted that without further experimental testing and validation in the highlighted compositional space, it cannot be confidently stated which model is a true and generalizable model. However, such subtle nuances and variations are observed through known and unknown underlying physical parameters can only be captured by a multi-source modeling data fusion framework built through LVGP. Furthermore, the current research study demonstrates that using LVGP as a multi-source model can be advantageous, especially when low amount of information exists about the material system of interest as the model provides smoother response surfaces closer to real physical material behavior, and accurate predictions that can be significantly beneficial for downstream tasks. 

\section{Conclusion}
AI and ML frameworks have had a strong impact in the science and engineering communities. On the other hand, with the ever-rising application of data-driven surrogates build on multiple sources of data that it is gathered either experimentally or computationally, not much attention has been paid to the quality, completeness, and comprehensiveness of the excluded and unknown underlying parameters involved in the data generation process. These parameters vary from differences in manufacturing process, testing technique, and more. Through the research done in this paper, an interpretable multi-source data fusion framework based on Latent Variable Gaussian Process (LVGP) is proposed and extensively studied. Through LVGP models, a categorical variable was introduced to represent sources of information during model training and predictions. A low-dimensional latent space is then learned to incorporate the categorical source variable into multi-source data fusion. Through latent variable representation, both the excluded and underlying unknown physical parameter variations among the different sources are captured and introduced into the modelling, making the multi-source ML model interpretable and aware of the differences and similarities between the information sources. The inclusion of the data sources not only leads to more accurate models but also enables the study of source specific response surfaces, material optimization, and sensitivity analysis. Compared to using single source data for ML, the proposed data fusion method using multi-source information can provide a better prediction for sparse-data problems by taking advantage of the information of correlations and relationships among different data sources. Additionally, a novel dissimilarity metric is also defined through the latent variables of the LVGP models that can help understand the aggregated differences and similarities in the information sources. The proposed approach is applied on and analyzed through two mathematical (representative parabola and 2D Ackley function) and two materials design (design of FeCrAl alloys and design of SmCoFe alloys) case studies.

Although the demonstrated method is focused on the material science domain, the proposed approach of leveraging LVGP for data fusion modeling can be applicable to other several domains of engineering where multiple source scenarios are observed frequently. The definition of a source, set of data points or a data generating process with different mapping behavior between the fixed set of inputs and outputs,  is what permeates the approach to domains. This definition of a source become inclusive of scenarios where sources are differentiated by fidelity to evaluate a design (e.g. potential flow solvers, Reynolds-Averaged Navier-Stokes and Large Eddy Simulation, for computational fluid dynamics applications, static and dynamic analysis for structural design), by differences in operational history of manufacturing systems (e.g., milling machine, binder jet machine) leading to machine to machine variations, and by differences in health of mechanical system (e.g., jet engines, gear systems) leading to performance variations. For these applications, the nuances between the sources does not need to be quantified directly and can be simply tagged as categorical variable into the LVGP model where the latent variables of the LVGP can internally learn the differences in sources. With the quantified uncertainty from LVGP for the specified information source, the approach can further guide adaptive sampling or active learning when allocating resources for data collection. As such, a novel application of the framework would involve performing materials design and development through autonomous experimentation studies where multiple sources of experimental data are gathered to guide the study. It is envisioned that the proposed multi-source framework is not limited to experimental or computational information sources and can be further extended to mixture of experimental and computational sources in numerous engineering and science applications.

Finally, a key challenge in the presented method lies when there are high number of sources present, potentially more than 15 sources. In that case, the LVGP model might struggle to converge to a final latent space due to high number of parameter (latent variables) estimations during model fitting.  However, it was previously shown that even when a large number of categorical variable options are present, further design optimization campaign works efficiently to identify top designs for metal-organic framework materials \cite{comlek2023rapid} . Due to time limitations for training GP models with the presence of large datasets, a future research direction for multi-source modeling through LVGP involves modeling through sparse-variational LVGP \cite{wang2022scalable}. Furthermore, as each information source could also contain different input parametrizations or spaces (heterogeneous inputs), extending LVGP to handle heterogeneous cases can benefit the numerous engineering applications \cite{comlek2024heterogenous}. 

\clearpage
\section*{Data Availability}
The MATLAB and python code for LVGP used in this study can be found at https://github.com/ideal-nu/MOF-LVGP-MOBBO \cite{comlek2023rapid} and https://github.com/syerramilli/lvgp-bayes \cite{yerramilli2023fully}.

\section*{Acknowledgments}
Northwestern faculty and student are grateful for the sponsorship from NSF 2219489 and the Army Research Laboratory under Cooperative Agreement Number W911NF-22-0121. The views and conclusions contained in this document are those of the authors and should not be interpreted as representing the official policies, either expressed or implied, of the Army Research Laboratory of the U.S. Government. The U.S. Government is authorized to reproduce and distribute reprints for Government purposes notwithstanding any copyright notation herein. This work is supported by the US Department of Energy, National Nuclear Security Administration, under award number DE-NE0009047. This material is based upon work supported by the U.S. Department of Energy’s Office of Energy Efficiency and Renewable Energy (EERE) under the Advanced Manufacturing Office, Award Number DE-AC0206H11357. The views expressed herein do not necessarily represent the views of the U.S. Department of Energy or the United States Government. 
\clearpage

\printbibliography

@article{ravi2022data,
  title={Data-driven modeling of multiaxial fatigue in frequency domain},
  author={Ravi, S.K. and Dong, P. and Wei, Z.},
  journal={Marine Structures},
  volume={84},
  pages={103201},
  year={2022},
  publisher={Elsevier}
}

@article{roy2023understanding,
  title={Understanding oxidation of Fe-Cr-Al alloys through explainable artificial intelligence},
  author={Roy, I. and Feng, B, and Roychowdhury, S. and Ravi, S.K., and Umretiya, R.V. and Reynolds, C. and Ghosh, S. and Rebak, R.B. and Hoffman, A.},
  journal={MRS communications},
  pages={1--7},
  year={2023},
  publisher={Springer}
}

@article{roy2023data,
  title={Data-driven predictive modeling of FeCrAl oxidation},
  author={Roy, I. and Roychowdhury, S. and Feng, B. and Ravi, S.K. and Ghosh, S. and Umretiya, R. and Rebak, R.B. and Ruscitto, D.M. and Gupta, V. and Hoffman, A.},
  journal={Materials Letters: X},
  pages={100183},
  year={2023},
  publisher={Elsevier}
}

@inproceedings{ravi2023uncertainty,
  title={On Uncertainty Quantification in Materials Modeling and Discovery: Applications of GE's BHM and IDACE},
  author={Ravi, S.K. and Bhaduri, A. and Amer, A. and Ghosh, S. and Wang, L. and Hoffman, A. and Umretiya, R. and Roy, I. and Rebak, R. and Dheeradhada, V.S. and others},
  booktitle={AIAA SCITECH 2023 Forum}
}

@article{williams2006gaussian,
  title={Gaussian processes for machine learning},
  author={Williams, Christopher K and Rasmussen, Carl Edward},
  journal={the MIT Press},
  volume={2},
  number={3},
  pages={4},
  year={2006}
}

@article{roy2023optimizing,
  title={Optimizing chemistry for designing oxidation resistant FeCrAl alloys},
  author={Roy, Indranil and Abouelella, Hamdy and Umretiya, Rajnikant V and Roychowdhury, Subhrajit and Feng, Bojun and Ravi, Sandipp Krishnan and Ghosh, Sayan and Rebak, Raul B and Ruscitto, Daniel M and Gupta, Vipul and others},
  journal={MRS Advances},
  volume={8},
  number={1},
  pages={21--26},
  year={2023},
  publisher={Springer}
}

@article{ravi2023elucidating,
  title={Elucidating precipitation in FeCrAl alloys through explainable AI: A case study},
  author={Ravi, Sandipp Krishnan and Roy, Indranil and Roychowdhury, Subhrajit and Feng, Bojun and Ghosh, Sayan and Reynolds, Christopher and Umretiya, Rajnikant V and Rebak, Raul B and Hoffman, Andrew K},
  journal={Computational Materials Science},
  volume={230},
  pages={112440},
  year={2023},
  publisher={Elsevier}
}

@article{wang2022scalable,
  title={Scalable gaussian processes for data-driven design using big data with categorical factors},
  author={Wang, Liwei and Yerramilli, Suraj and Iyer, Akshay and Apley, Daniel and Zhu, Ping and Chen, Wei},
  journal={Journal of Mechanical Design},
  volume={144},
  number={2},
  pages={021703},
  year={2022},
  publisher={American Society of Mechanical Engineers}
}

@article{zhang2020latent,
  title={A latent variable approach to Gaussian process modeling with qualitative and quantitative factors},
  author={Zhang, Yichi and Tao, Siyu and Chen, Wei and Apley, Daniel W},
  journal={Technometrics},
  volume={62},
  number={3},
  pages={291--302},
  year={2020},
  publisher={Taylor \& Francis}
}

@article{comlek2023mixed,
  title={Mixed-Variable Global Sensitivity Analysis for Knowledge Discovery And Efficient Combinatorial Materials Design (IDETC2023-110756)},
  author={Comlek, Yigitcan and Wang, Liwei and Chen, Wei},
  journal={Journal of Mechanical Design},
  pages={1--31},
  year={2023}
}

@article{comlek2023rapid,
  title={Rapid design of top-performing metal-organic frameworks with qualitative representations of building blocks},
  author={Comlek, Yigitcan and Pham, Thang Duc and Snurr, Randall Q and Chen, Wei},
  journal={npj Computational Materials},
  volume={9},
  number={1},
  pages={170},
  year={2023},
  publisher={Nature Publishing Group UK London}
}

@article{comlek2024heterogenous,
  title={Heterogenous Multi-Source Data Fusion Through Input Mapping and Latent Variable Gaussian Process},
  author={Comlek, Yigitcan and Ravi, Sandipp Krishnan and Pandita, Piyush and Ghosh, Sayan and Wang, Liping and Chen, Wei},
  journal={arXiv preprint arXiv:2407.11268},
  year={2024}
}

@article{himanen2019data,
  title={Data-driven materials science: status, challenges, and perspectives},
  author={Himanen, Lauri and Geurts, Amber and Foster, Adam Stuart and Rinke, Patrick},
  journal={Advanced Science},
  volume={6},
  number={21},
  pages={1900808},
  year={2019},
  publisher={Wiley Online Library}
}

@inproceedings{sambasivan2021everyone,
  title={“Everyone wants to do the model work, not the data work”: Data Cascades in High-Stakes AI},
  author={Sambasivan, Nithya and Kapania, Shivani and Highfill, Hannah and Akrong, Diana and Paritosh, Praveen and Aroyo, Lora M},
  booktitle={proceedings of the 2021 CHI Conference on Human Factors in Computing Systems},
  pages={1--15},
  year={2021}
}

@article{liang2022advances,
  title={Advances, challenges and opportunities in creating data for trustworthy AI},
  author={Liang, Weixin and Tadesse, Girmaw Abebe and Ho, Daniel and Fei-Fei, L and Zaharia, Matei and Zhang, Ce and Zou, James},
  journal={Nature Machine Intelligence},
  volume={4},
  number={8},
  pages={669--677},
  year={2022},
  publisher={Nature Publishing Group UK London}
}

@article{blaiszik2016materials,
  title={The materials data facility: data services to advance materials science research},
  author={Blaiszik, Ben and Chard, Kyle and Pruyne, Jim and Ananthakrishnan, Rachana and Tuecke, Steven and Foster, Ian},
  journal={Jom},
  volume={68},
  number={8},
  pages={2045--2052},
  year={2016},
  publisher={Springer}
}

@article{puchala2016materials,
  title={The materials commons: a collaboration platform and information repository for the global materials community},
  author={Puchala, Brian and Tarcea, Glenn and Marquis, Emmanuelle A and Hedstrom, Margaret and Jagadish, HV and Allison, John E},
  journal={Jom},
  volume={68},
  pages={2035--2044},
  year={2016},
  publisher={Springer}
}

@article{gong2022repository,
  title={A repository for the publication and sharing of heterogeneous materials data},
  author={Gong, Haiyan and He, Jie and Zhang, Xiaotong and Duan, Lei and Tian, Ziqi and Zhao, Wei and Gong, Fuzhou and Liu, Tong and Wang, Zongguo and Zhao, Haifeng and others},
  journal={Scientific Data},
  volume={9},
  number={1},
  pages={787},
  year={2022},
  publisher={Nature Publishing Group UK London}
}

@article{kirklin2015open,
  title={The Open Quantum Materials Database (OQMD): assessing the accuracy of DFT formation energies},
  author={Kirklin, Scott and Saal, James E and Meredig, Bryce and Thompson, Alex and Doak, Jeff W and Aykol, Muratahan and R{\"u}hl, Stephan and Wolverton, Chris},
  journal={npj Computational Materials},
  volume={1},
  number={1},
  pages={1--15},
  year={2015},
  publisher={Nature Publishing Group}
}

@article{jain2013commentary,
  title={Commentary: The Materials Project: A materials genome approach to accelerating materials innovation},
  author={Jain, Anubhav and Ong, Shyue Ping and Hautier, Geoffroy and Chen, Wei and Richards, William Davidson and Dacek, Stephen and Cholia, Shreyas and Gunter, Dan and Skinner, David and Ceder, Gerbrand and others},
  journal={APL materials},
  volume={1},
  number={1},
  year={2013},
  publisher={AIP Publishing}
}

@article{nti2022applications,
  title={Applications of artificial intelligence in engineering and manufacturing: A systematic review},
  author={Nti, Isaac Kofi and Adekoya, Adebayo Felix and Weyori, Benjamin Asubam and Nyarko-Boateng, Owusu},
  journal={Journal of Intelligent Manufacturing},
  volume={33},
  number={6},
  pages={1581--1601},
  year={2022},
  publisher={Springer}
}

@article{iyer2020data,
  title={Data centric nanocomposites design via mixed-variable Bayesian optimization},
  author={Iyer, Akshay and Zhang, Yichi and Prasad, Aditya and Gupta, Praveen and Tao, Siyu and Wang, Yixing and Prabhune, Prajakta and Schadler, Linda S and Brinson, L Catherine and Chen, Wei},
  journal={Molecular Systems Design \& Engineering},
  volume={5},
  number={8},
  pages={1376--1390},
  year={2020},
  publisher={Royal Society of Chemistry}
}

@article{wang2020featureless,
  title={Featureless adaptive optimization accelerates functional electronic materials design},
  author={Wang, Yiqun and Iyer, Akshay and Chen, Wei and Rondinelli, James M},
  journal={Applied Physics Reviews},
  volume={7},
  number={4},
  year={2020},
  publisher={AIP Publishing}
}

@article{zhang2020bayesian,
  title={Bayesian optimization for materials design with mixed quantitative and qualitative variables},
  author={Zhang, Yichi and Apley, Daniel W and Chen, Wei},
  journal={Scientific reports},
  volume={10},
  number={1},
  pages={4924},
  year={2020},
  publisher={Nature Publishing Group UK London}
}

@article{kobayashi2010mapping,
  title={Mapping of 475 C embrittlement in ferritic Fe--Cr--Al alloys},
  author={Kobayashi, Satoru and Takasugi, Takayuki},
  journal={Scripta Materialia},
  volume={63},
  number={11},
  pages={1104--1107},
  year={2010},
  publisher={Elsevier}
}

@article{han2016effect,
  title={Effect of Cr/Al contents on the 475{\textordmasculine}C age-hardening in oxide dispersion strengthened ferritic steels},
  author={Han, Wentuo and Yabuuchi, Kiyohiro and Kimura, Akihiko and Ukai, Shigeharu and Oono, Naoko and Kaito, Takeji and Torimaru, Tadahiko and Hayashi, Shigenari},
  journal={Nuclear Materials and Energy},
  volume={9},
  pages={610--615},
  year={2016},
  publisher={Elsevier}
}

@article{kim2019400,
  title={400 C aging embrittlement of FeCrAl alloys: Microstructure and fracture behavior},
  author={Kim, Hyunmyung and Subramanian, Gokul Obulan and Kim, Chaewon and Jang, Hun and Jang, Changheui},
  journal={Materials Science and Engineering: A},
  volume={743},
  pages={159--167},
  year={2019},
  publisher={Elsevier}
}

@article{ejenstam2015microstructural,
  title={Microstructural stability of Fe--Cr--Al alloys at 450--550 C},
  author={Ejenstam, Jesper and Thuvander, Mattias and Olsson, P{\"a}r and Rave, Fernando and Szakalos, Peter},
  journal={Journal of Nuclear Materials},
  volume={457},
  pages={291--297},
  year={2015},
  publisher={Elsevier}
}

@article{yang2020aluminum,
  title={Aluminum suppression of $\alpha'$ precipitate in model Fe--Cr--Al alloys during long-term aging at 475 C},
  author={Yang, Z and Wang, ZX and Xia, CH and Ouyang, MH and Peng, JC and Zhang, HW and Xiao, XS},
  journal={Materials Science and Engineering: A},
  volume={772},
  pages={138714},
  year={2020},
  publisher={Elsevier}
}

@article{capdevila2012phase,
  title={Phase separation kinetics in a Fe--Cr--Al alloy},
  author={Capdevila, Carlos and Miller, Michael K and Chao, Jes{\'u}s},
  journal={Acta materialia},
  volume={60},
  number={12},
  pages={4673--4684},
  year={2012},
  publisher={Elsevier}
}

@article{capdevila2008phase,
  title={Phase separation in PM 2000™ Fe-base ODS alloy: Experimental study at the atomic level},
  author={Capdevila, Carlos and Miller, Michael K and Russell, Kaye F and Chao, Jes{\'u}s and Gonz{\'a}lez-Carrasco, Jos{\'e} Luis},
  journal={Materials Science and Engineering: A},
  volume={490},
  number={1-2},
  pages={277--288},
  year={2008},
  publisher={Elsevier}
}

@article{pinkas2015sensitivity,
  title={Sensitivity of thermo-electric power measurements to $\alpha$--$\alpha'$ phase separation in Cr-rich oxide dispersion strengthened steels},
  author={Pinkas, Malki and Foxman, Zvi and Froumin, Nataly and H{\"a}hner, Peter and Meshi, Louisa},
  journal={Journal of Materials Science},
  volume={50},
  pages={4629--4635},
  year={2015},
  publisher={Springer}
}

@article{senno1974permanent,
  title={Permanent-magnet properties of Sm-Ce-Co-Fe-Cu alloys with compositions between 1-5 and 2-17},
  author={Senno, Harufumi and Tawara, YOSHIO},
  journal={IEEE Transactions on Magnetics},
  volume={10},
  number={2},
  pages={313--317},
  year={1974},
  publisher={IEEE}
}

@patent{US20150262740A1,
  title = {Rare earth-cobalt permanent magnet},
  author = {Teruhiko Fujiwara, Hiroaki Machida, Hideyuki Yoshikawa},
  year = {2015},
  note = {Tokin Coorporation},
  url = {https://patents.google.com/patent/US20150262740A1/en?oq=U.S.+Patent+Application+No.+14\%2f643\%2c875.},
}

@book{ackley2012connectionist,
  title={A connectionist machine for genetic hillclimbing},
  author={Ackley, David},
  volume={28},
  year={2012},
  publisher={Springer science \& business media}
}

@article{batra2019multifidelity,
  title={Multifidelity information fusion with machine learning: A case study of dopant formation energies in hafnia},
  author={Batra, Rohit and Pilania, Ghanshyam and Uberuaga, Blas P and Ramprasad, Rampi},
  journal={ACS applied materials \& interfaces},
  volume={11},
  number={28},
  pages={24906--24918},
  year={2019},
  publisher={ACS Publications}
}

@article{pilania2017multi,
  title={Multi-fidelity machine learning models for accurate bandgap predictions of solids},
  author={Pilania, Ghanshyam and Gubernatis, James E and Lookman, Turab},
  journal={Computational Materials Science},
  volume={129},
  pages={156--163},
  year={2017},
  publisher={Elsevier}
}

@article{cui2018current,
  title={Current progress and future challenges in rare-earth-free permanent magnets},
  author={Cui, Jun and Kramer, Matthew and Zhou, Lin and Liu, Fei and Gabay, Alexander and Hadjipanayis, George and Balasubramanian, Balamurugan and Sellmyer, David},
  journal={Acta Materialia},
  volume={158},
  pages={118--137},
  year={2018},
  publisher={Elsevier}
}

@article{chikhalikar2022effect,
  title={Effect of aluminum on the FeCr (Al) alloy oxidation resistance in steam environment at low temperature (400 C) and high temperature (1200 C)},
  author={Chikhalikar, Atharva and Roy, Indranil and Abouelella, Hamdy and Umretiya, Rajnikant and Hoffman, Andrew and Larsen, Mike and Rebak, Raul B},
  journal={Corrosion Science},
  volume={209},
  pages={110765},
  year={2022},
  publisher={Elsevier}
}

@article{field2018precipitation,
  title={Precipitation of $\alpha'$ in neutron irradiated commercial FeCrAl alloys},
  author={Field, Kevin G and Littrell, Kenneth C and Briggs, Samuel A},
  journal={Scripta Materialia},
  volume={142},
  pages={41--45},
  year={2018},
  publisher={Elsevier}
}

@article{capdevila2008aluminum,
  title={Aluminum partitioning during phase separation in Fe--20\% Cr--6\% Al ODS alloy},
  author={Capdevila, C and Miller, Michael K and Russell, Kaye F},
  journal={Journal of materials science},
  volume={43},
  pages={3889--3893},
  year={2008},
  publisher={Springer}
}

@article{li2013effect,
  title={The effect of Al on the 475 C embrittlement of Fe--Cr alloys},
  author={Li, Wei and Lu, Song and Hu, Qing-Miao and Mao, Huahai and Johansson, B{\"o}rje and Vitos, Levente},
  journal={Computational materials science},
  volume={74},
  pages={101--106},
  year={2013},
  publisher={Elsevier}
}

@article{hoffman2021effects,
  title={Effects of Al on Alpha Prime Formation in FeCrAl Alloys},
  author={Hoffman, AK and Nag, S and Chen, C and Rebak, RB and Lutz, DR and Jiang, C},
  journal={Proceedings of the TopFuel},
  year={2021}
}

@article{meng2020survey,
  title={A survey on machine learning for data fusion},
  author={Meng, Tong and Jing, Xuyang and Yan, Zheng and Pedrycz, Witold},
  journal={Information Fusion},
  volume={57},
  pages={115--129},
  year={2020},
  publisher={Elsevier}
}

@article{bleiholder2009data,
  title={Data fusion},
  author={Bleiholder, Jens and Naumann, Felix},
  journal={ACM computing surveys (CSUR)},
  volume={41},
  number={1},
  pages={1--41},
  year={2009},
  publisher={ACM New York, NY, USA}
}

@article{zhou2019information,
  title={Information fusion for multi-source material data: Progress and challenges},
  author={Zhou, Jingren and Hong, Xin and Jin, Peiquan},
  journal={Applied Sciences},
  volume={9},
  number={17},
  pages={3473},
  year={2019},
  publisher={MDPI}
}

@article{CHEN2024116773,
title = {A Latent Variable Approach for Non-Hierarchical Multi-Fidelity Adaptive Sampling},
journal = {Computer Methods in Applied Mechanics and Engineering},
volume = {421},
pages = {116773},
year = {2024},
issn = {0045-7825},
author = {Yi-Ping Chen and Liwei Wang and Yigitcan Comlek and Wei Chen},
}

@book{ackley2,
    author = {Bäck, Thomas},
    title = {Evolutionary Algorithms in Theory and Practice: Evolution Strategies, Evolutionary Programming, Genetic Algorithms},
    publisher = {Oxford University Press},
    year = {1996},
    month = {02},
    isbn = {9780195099713},
}

@article{yerramilli2023fully,
  title={Fully bayesian inference for latent variable gaussian process models},
  author={Yerramilli, Suraj and Iyer, Akshay and Chen, Wei and Apley, Daniel W},
  journal={SIAM/ASA Journal on Uncertainty Quantification},
  volume={11},
  number={4},
  pages={1357--1381},
  year={2023},
  publisher={SIAM}
}

@article{cook2005sufficient,
  title={Sufficient dimension reduction via inverse regression: A minimum discrepancy approach},
  author={Cook, R Dennis and Ni, Liqiang},
  journal={Journal of the American Statistical Association},
  volume={100},
  number={470},
  pages={410--428},
  year={2005},
  publisher={Taylor \& Francis}
}

@article{li1991sliced,
  title={Sliced inverse regression for dimension reduction},
  author={Li, Ker-Chau},
  journal={Journal of the American Statistical Association},
  volume={86},
  number={414},
  pages={316--327},
  year={1991},
  publisher={Taylor \& Francis}
}

@article{wang2024deep,
  title={A deep learning interpretable model for river dissolved oxygen multi-step and interval prediction based on multi-source data fusion},
  author={Wang, Zhaocai and Wang, Qingyu and Liu, Zhixiang and Wu, Tunhua},
  journal={Journal of Hydrology},
  volume={629},
  pages={130637},
  year={2024},
  publisher={Elsevier}
}

@article{yao2023ensemble,
  title={An ensemble CNN-LSTM and GRU adaptive weighting model based improved sparrow search algorithm for predicting runoff using historical meteorological and runoff data as input},
  author={Yao, Zhiyuan and Wang, Zhaocai and Wang, Dangwei and Wu, Junhao and Chen, Lingxuan},
  journal={Journal of Hydrology},
  volume={625},
  pages={129977},
  year={2023},
  publisher={Elsevier}
}

@article{zhang2021multi,
  title={Multi-source information fusion based on rough set theory: A review},
  author={Zhang, Pengfei and Li, Tianrui and Wang, Guoqiang and Luo, Chuan and Chen, Hongmei and Zhang, Junbo and Wang, Dexian and Yu, Zeng},
  journal={Information Fusion},
  volume={68},
  pages={85--117},
  year={2021},
  publisher={Elsevier}
}

@article{yousefpour2024gp+,
  title={GP+: a python library for kernel-based learning via Gaussian Processes},
  author={Yousefpour, Amin and Foumani, Zahra Zanjani and Shishehbor, Mehdi and Mora, Carlos and Bostanabad, Ramin},
  journal={Advances in Engineering Software},
  volume={195},
  pages={103686},
  year={2024},
  publisher={Elsevier}
}

@article{eweis2022data,
  title={Data fusion with latent map Gaussian processes},
  author={Eweis-Labolle, Jonathan Tammer and Oune, Nicholas and Bostanabad, Ramin},
  journal={Journal of Mechanical Design},
  volume={144},
  number={9},
  pages={091703},
  year={2022},
  publisher={American Society of Mechanical Engineers}
}

@article{foumani2023multi,
  title={Multi-fidelity cost-aware Bayesian optimization},
  author={Foumani, Zahra Zanjani and Shishehbor, Mehdi and Yousefpour, Amin and Bostanabad, Ramin},
  journal={Computer Methods in Applied Mechanics and Engineering},
  volume={407},
  pages={115937},
  year={2023},
  publisher={Elsevier}
}

@article{zanjani2024safeguarding,
  title={Safeguarding Multi-fidelity Bayesian Optimization Against Large Model Form Errors and Heterogeneous Noise},
  author={Zanjani Foumani, Zahra and Yousefpour, Amin and Shishehbor, Mehdi and Bostanabad, Ramin},
  journal={Journal of Mechanical Design},
  volume={146},
  number={6},
  year={2024},
  publisher={American Society of Mechanical Engineers Digital Collection}
}

@article{wang2021data,
  title={Data-driven topology optimization with multiclass microstructures using latent variable Gaussian process},
  author={Wang, Liwei and Tao, Siyu and Zhu, Ping and Chen, Wei},
  journal={Journal of Mechanical Design},
  volume={143},
  number={3},
  pages={031708},
  year={2021},
  publisher={American Society of Mechanical Engineers}
}

\clearpage
\setcounter{equation}{0}
\setcounter{figure}{0}
\setcounter{table}{0}

\section*{Supplementary Information}

\subsection*{Supplementary Figures}
\begin{figure}[hbt!]
\centering
\includegraphics[width=.8\textwidth]{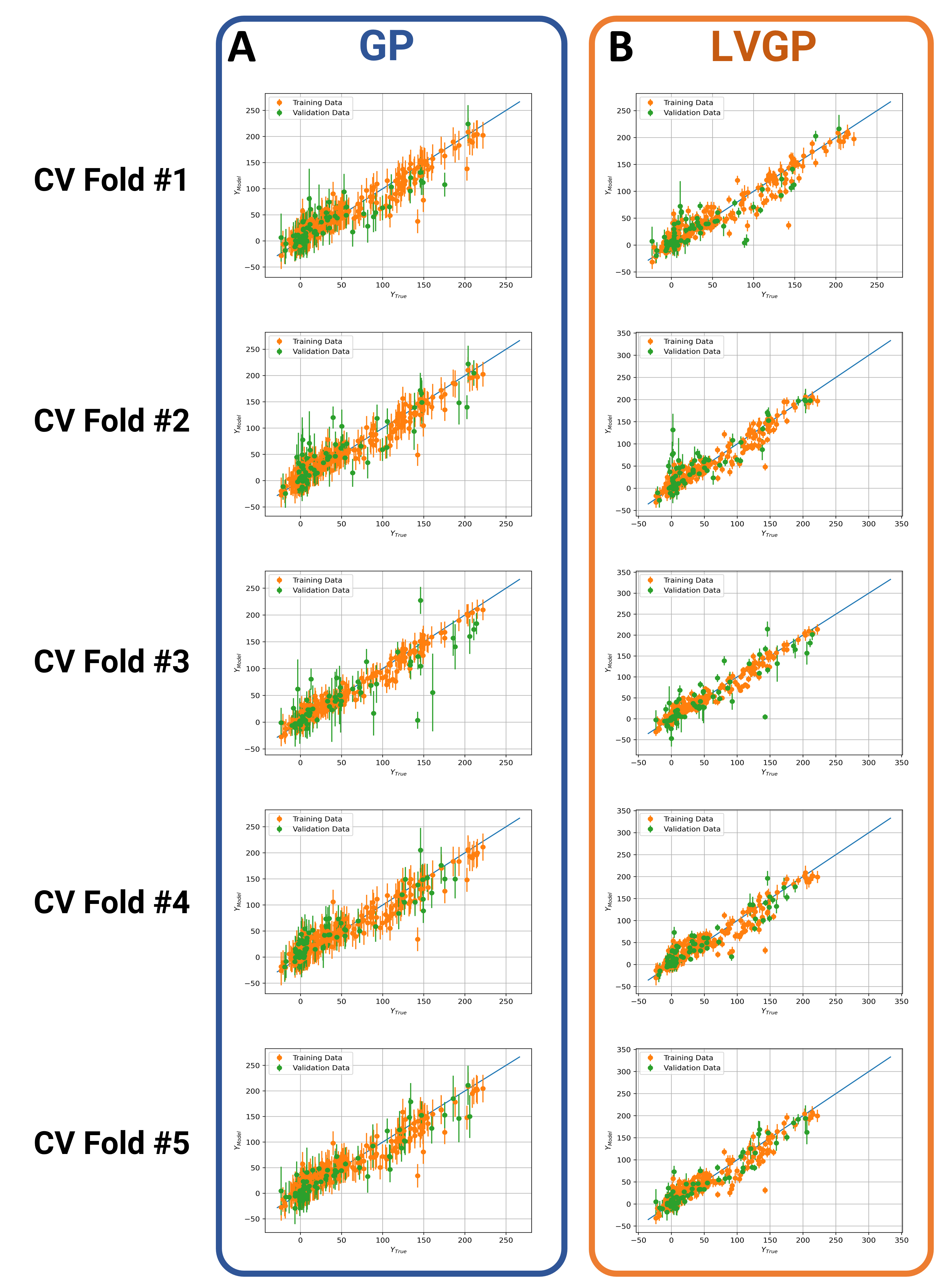}
\caption{Cross Validation performance on FeCrAl Alloy dataset using (a) GP and (b) LVGP models. The x and y axis show the true and predicted hardness change (HV), respectively, and error bars demonstrate the prediction uncertainty associated with each prediction}
\label{fig:fecral_cv}
\end{figure}

\begin{figure}[hbt!]
\centering
\includegraphics[width=.5\textwidth]{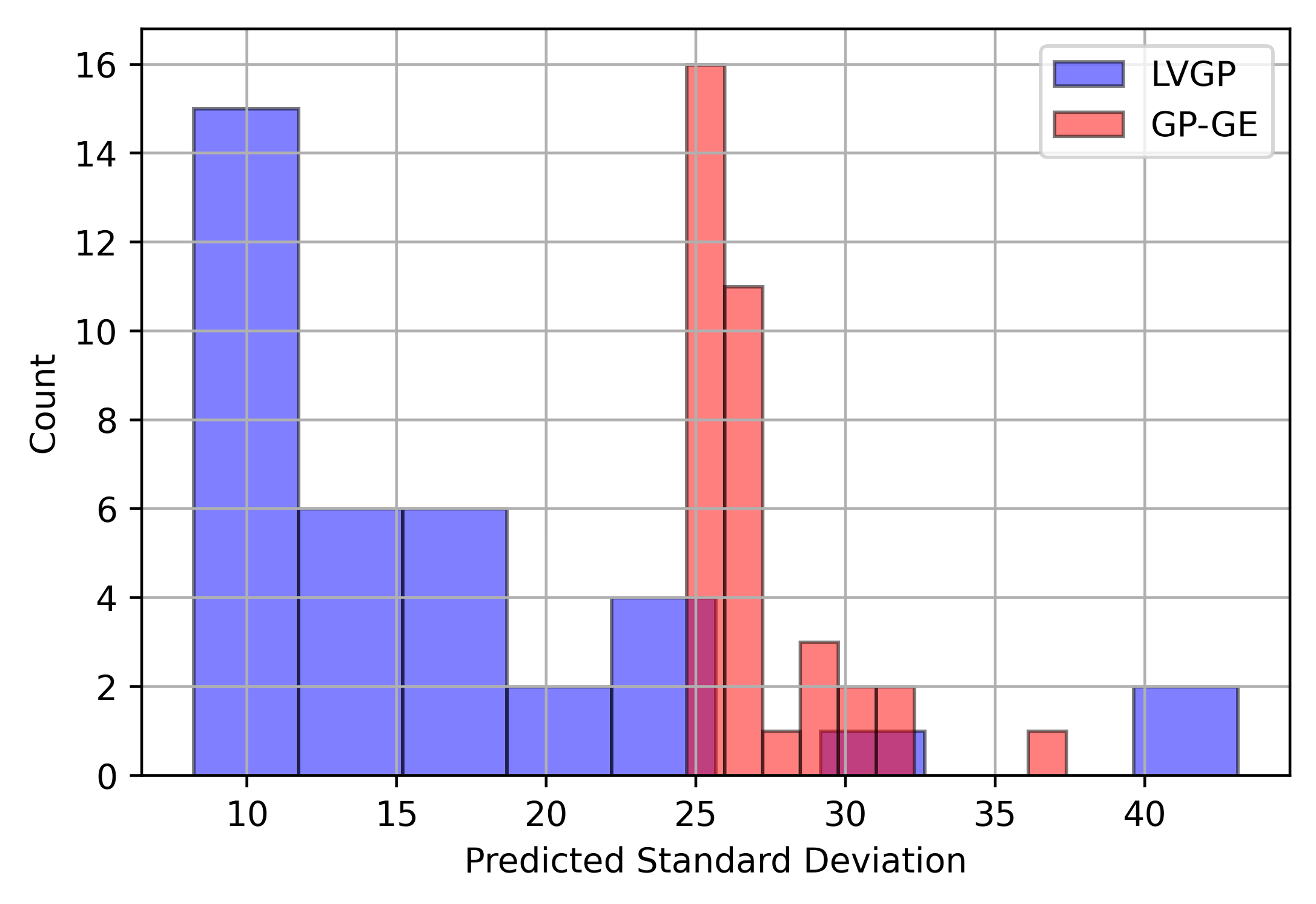}
\caption{Predicted uncertainty distribution provided by the LVGP and GP-GE models on the 36 sample GE Data testing set}
\label{fig:fecral_std_si}
\end{figure}

\begin{figure}[hbt!]
\centering
\includegraphics[width=.5\textwidth]{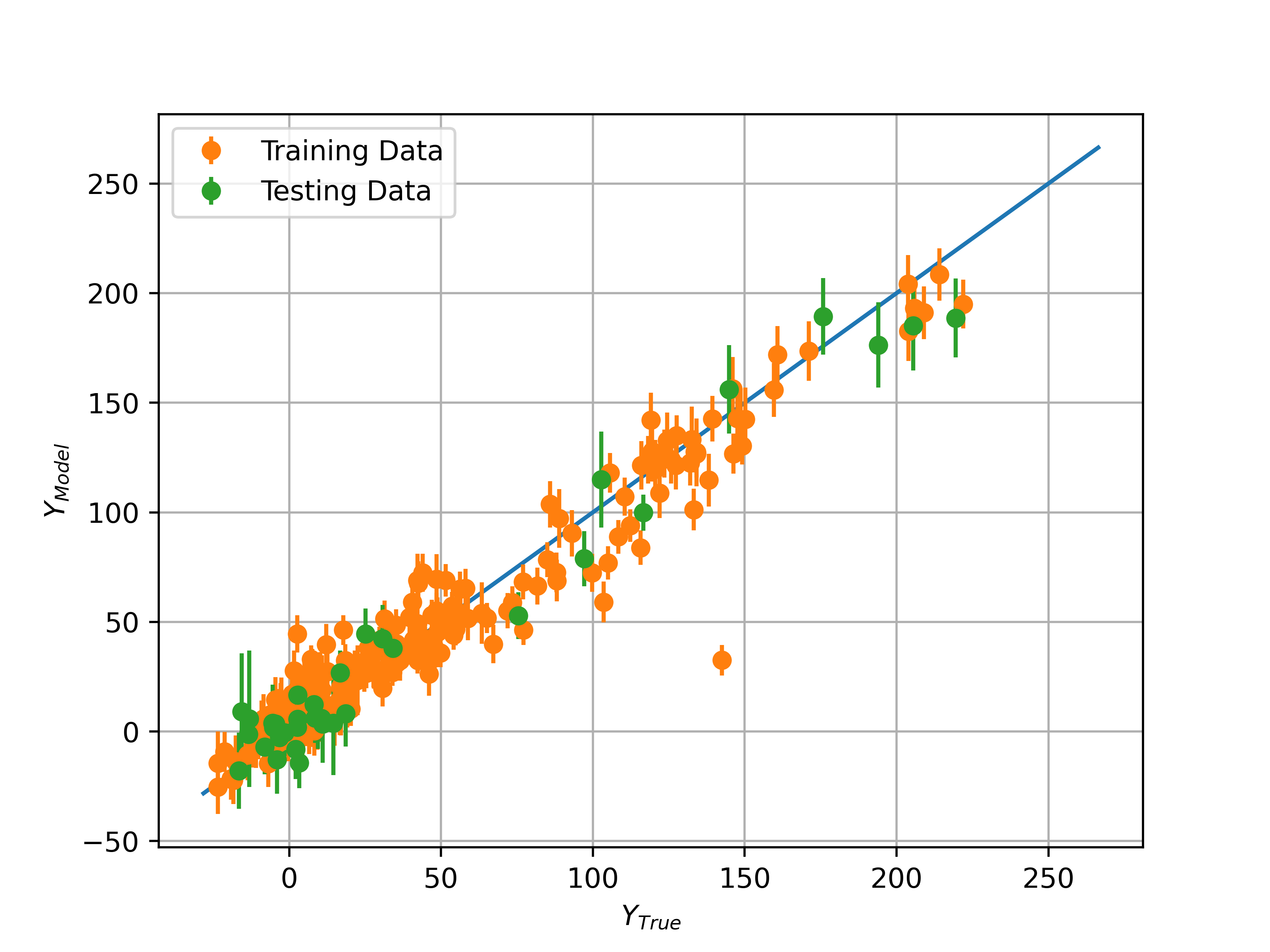}
\caption{Training and testing performance on FeCrAl Alloy dataset using the LVGP-Target model. The x and y axis show the true and predicted hardness change (HV), respectively, and error bars demonstrate the prediction uncertainty associated with each prediction}
\label{fig:fecral_gptarget}
\end{figure}

\begin{figure}[hbt!]
\centering
\includegraphics[width=.8\textwidth]{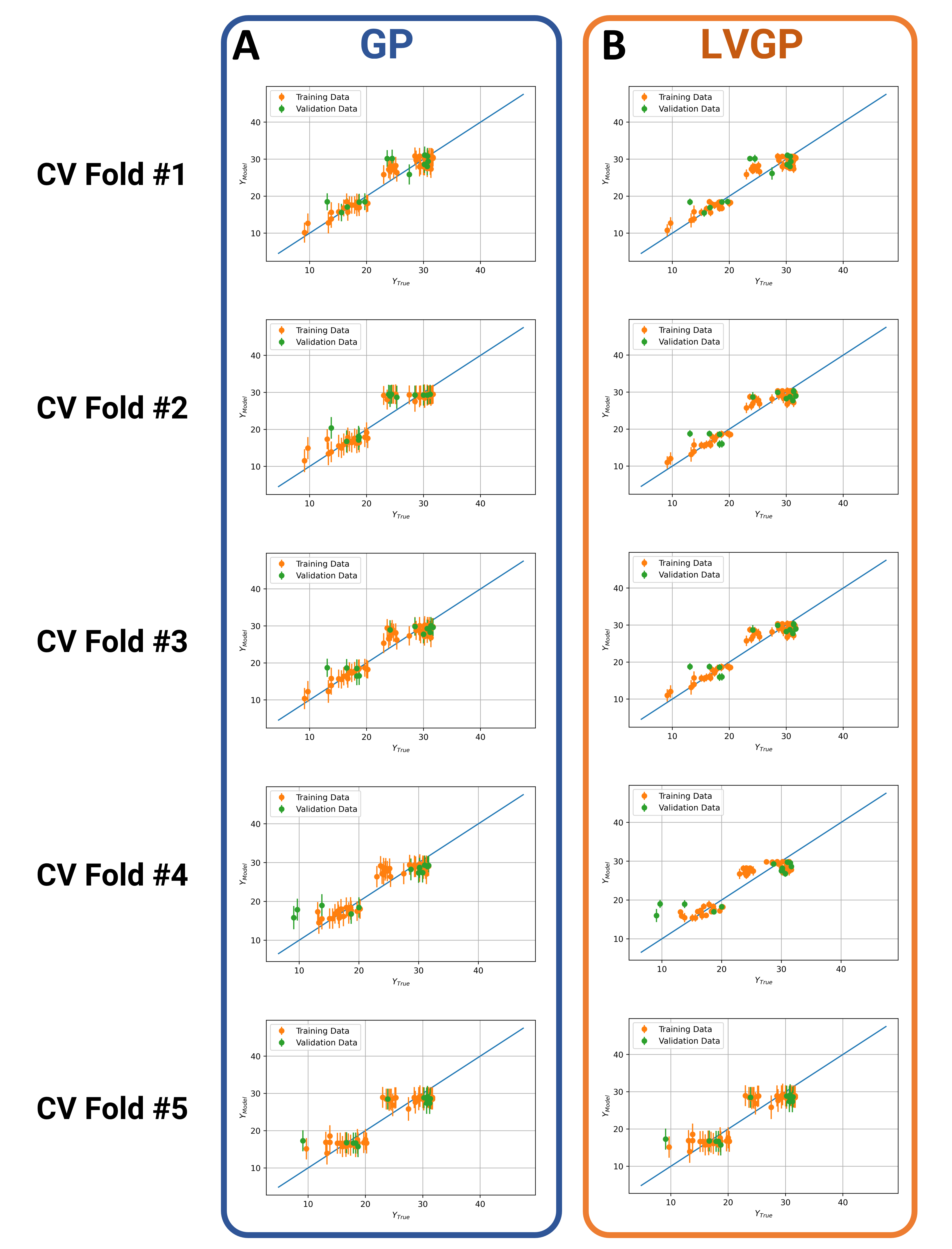}
\caption{Cross Validation performance on SmCoFe Alloy dataset using (a) GP and (b) LVGP models. The x and y axis show the true and predicted energy potential (EP), respectively, and error bars demonstrate the prediction uncertainity associated with each prediction}
\label{fig:smcofe_cv}
\end{figure}

\begin{figure}[hbt!]
\centering
\includegraphics[width=.5\textwidth]{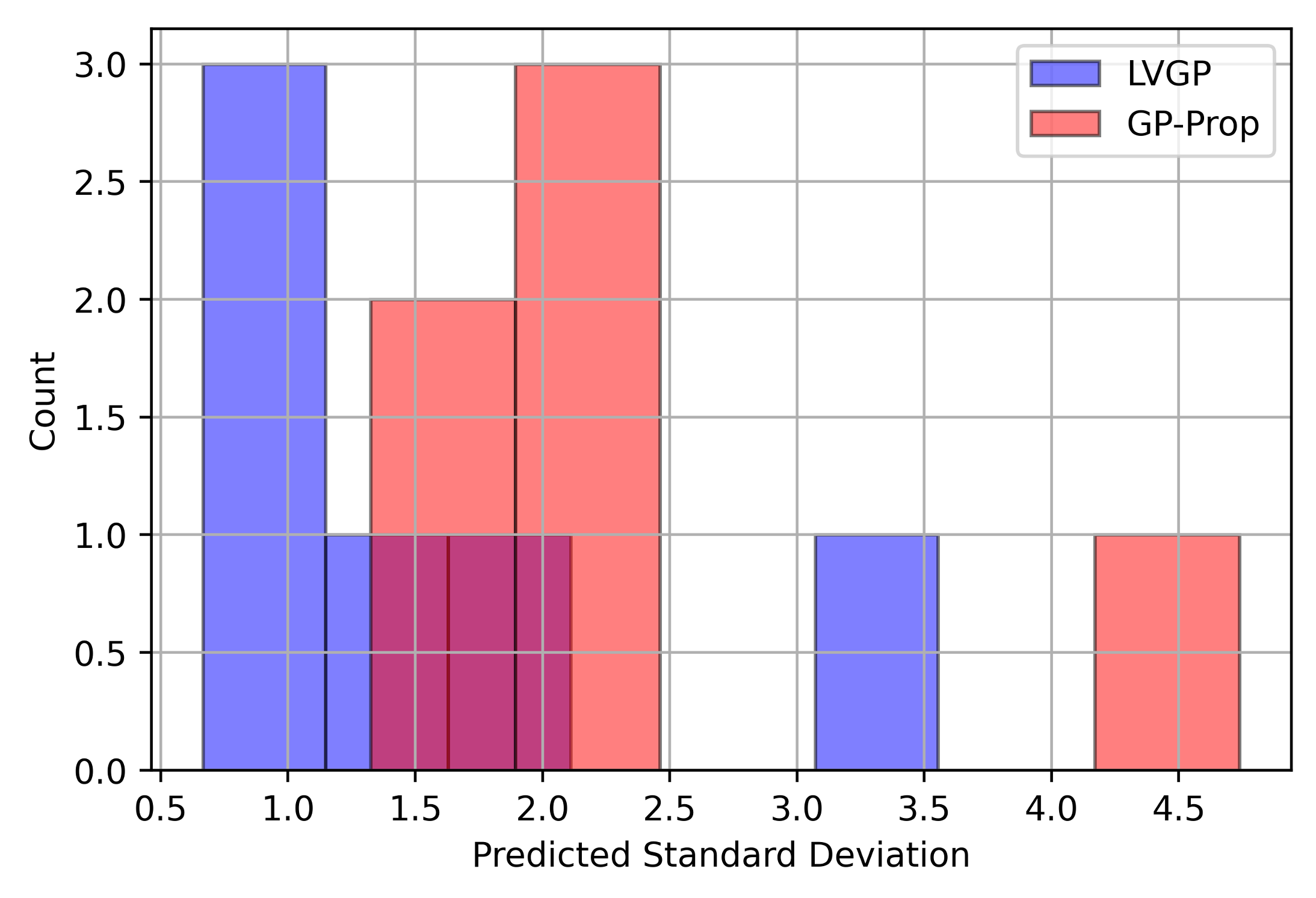}
\caption{Predicted uncertainty distribution provided by the LVGP and GP-Prof models on the 6 sample Prop of SmCeCoFeCu testing set}
\label{fig:smcofe_std_si}
\end{figure}

\clearpage

\subsection*{Supplementary Tables}
\begin{table*}[h]
\small
\centering
\caption{Sources of Information for the Parabola Problem}
\label{tab:parabola data}
\renewcommand{\arraystretch}{1.3}
\begin{tabular}{ccccccccc}
\toprule
Information Source & $a$ & $b$ & $x_{shift}$ & $y_{shift}$ & Number of Training Samples & Number of Testing Samples \\
\toprule
Ground Source & $1$ & $2$ & $0$ & $0$ & $2$ & $30$   \\
\\
Perturbed Source - 1 & $1$ & $2$ & $8$ & $0$ & $10$ & $30$   \\
\\
Perturbed Source - 2 & $1$ & $2$ & $0$ & $100$ & $10$ & $30$   \\
\\
Perturbed Source - 3 & $1$ & $2$ & $12$ & $120$ & $10$ & $30$   \\
\\
\toprule
\end{tabular}
\end{table*}

\begin{table*}[h]
\small
\centering
\caption{Sources of Information for the 2D Ackley Problem} 
\label{tab:ackley_data}
\renewcommand{\arraystretch}{1.3}
\begin{tabular}{p{3cm}p{7.5cm}p{2.5cm}p{2.5cm}ccc}
\toprule
Information Source & Function Formulation $(a = 20, b= 0.2, c=2)$ & Number of Training Samples & Number of Testing Samples \\
\toprule
Ground Source & $ z = -a*exp(-b(\sqrt{\frac{1}{2}(x^2-y^2)})) - exp(\frac{1}{2}(cos(cx)+cos(cy)))+a+exp(1)$ & $20$ & $100$   \\
\\
Perturbed Source - 1 & $z = -a*exp(-b(\sqrt{\frac{1}{2}(x^2-y^2)})) +10$ & $50$ & $100$   \\
\\
Perturbed Source - 2 & $z = exp(\frac{1}{2}(cos(cx)+cos(xy)))+5$ & $50$ & $100$   \\
\\
Perturbed Source - 3 & $z = \frac{1}{4}(-a*exp(-b(\sqrt{\frac{1}{2}(x^2-y^2)}))) - \frac{3}{4}(exp(\frac{1}{2}(cos(cx)+cos(cy))))+a+exp(1)$ & $50$ & $100$   \\
\\
\toprule
\end{tabular}
\end{table*}

\end{document}